\documentclass[preprint,11pt,a4paper]{elsarticle}

\usepackage[utf8]{inputenc}
\usepackage[T1]{fontenc}
\usepackage{lmodern}
\usepackage{microtype}
\usepackage[a4paper,margin=2.5cm]{geometry}
\usepackage{enumitem}
\setlist[enumerate]{leftmargin=1.6em,itemsep=0.35em,topsep=0.4em,parsep=0pt}
\usepackage[british]{babel}
\usepackage{amsmath,amssymb}
\usepackage{siunitx}
\DeclareSIUnit{\pp}{pp}
\DeclareSIUnit{\byte}{B}

\usepackage{booktabs}
\usepackage{array}
\usepackage{makecell}
\usepackage{graphicx}
\usepackage{caption}
\usepackage{xcolor}
\usepackage[hyphens]{url}
\usepackage{hyperref}
\hypersetup{colorlinks=true, linkcolor=black, citecolor=black, urlcolor=blue!60!black}

\usepackage[capitalise,noabbrev]{cleveref}

\usepackage[section]{placeins}

\newcommand{\rc}[1]{\textit{(#1)}}

\AtBeginDocument{\pdfpagewidth=\paperwidth \pdfpageheight=\paperheight}

\journal{Journal of Systems Architecture}
\makeatletter
\def\ps@pprintTitle{\let\@oddhead\@empty\let\@evenhead\@empty
  \def\@oddfoot{\centerline{\thepage}}\let\@evenfoot\@oddfoot}
\makeatother

\begin{document}

\begin{frontmatter}

\title{Does synthetic pre-training pay off on a microcontroller? Accuracy, latency and energy across
two MCUs}

\author[rmit,bits]{Achintya Yadav\corref{cor1}}
\ead{yadav.achintya2005@gmail.com}
\cortext[cor1]{Corresponding author.}

\affiliation[rmit]{department={School of Engineering,},
            organization={RMIT University},
            addressline={GPO Box 2476},
            city={Melbourne},
            postcode={3001},
            state={VIC},
            country={Australia}}

\affiliation[bits]{organization={Birla Institute of Technology and Science, Pilani},
            addressline={Hyderabad Campus, Jawahar Nagar, Kapra Mandal, Medchal District},
            city={Hyderabad},
            postcode={500078},
            state={Telangana},
            country={India}}
\begin{abstract}
Rendered images are cheap to label, but a microcontroller classifier must survive 8-bit integer
(INT8) quantisation and compute correctly on the chip. Five models, from MobileNetV2 at
\SI{224}{px} to a 61K-parameter convolutional neural network (CNN), were fine-tuned on a six-class scene dataset with and without
a synthetic stage, against thresholds fixed in advance, and deployed on an ESP32-S3 and an RP2040.
With full real data the stage gave no detectable gain. For the main deployable model the INT8
difference was \SI{+0.312}{} percentage points (pp), its interval below the \SI{+1.6}{\pp} gain recomputed from the
source study; quantisation did not detectably change it, nor did model size. With 10\% of the real
data the stage paid off for the CNN, by \SI{+3.438}{\pp} on one seed and \SI{+2.715}{\pp} over
three; its cause is open. On the ESP32-S3 the main model took \SI{152.7}{\milli\second}
and \SI{16.3}{\milli\joule} above idle per inference, and adding the stage changed neither beyond
preset bounds. The model's INT8 outputs matched TensorFlow Lite's reference kernels exactly, but
only after a memory overrun in a vendor kernel was patched. On the RP2040 the same model took
\SI{1.80}{\second} and \SI{27.3}{\milli\joule} above idle, about twelve times slower, and the
largest \SI{96}{px} model exceeded the board's flash-capacity limit. Neither the RP2040 nor the host
interpreter used for scoring reproduced the reference arithmetic. A synthetic stage is worth adding
only when real data are scarce, and on-device INT8 outputs should be checked against reference
kernels, not inferred from host scores.
\end{abstract}


\end{frontmatter}

\section{Introduction}\label{sec:intro}
A classifier that runs on a microcontroller is usually trained on a small set of labelled
photographs from its target setting, starting from weights pre-trained on
ImageNet~\cite{deng2009,russakovsky2015}. Rendered images are a tempting second source: a game
engine can produce labelled scenes in any quantity at almost no labelling cost, and computer-game
imagery has been used as training data since at least 2016~\cite{richter2016}. Whether a synthetic
pre-training stage still helps once the model has been shrunk, quantised to 8-bit integers and put
on the chip is a separate question, and building a synthetic corpus is only worth the time if the
answer is yes.

The study this paper re-examines is Bird et al.~\cite{bird2020}. They trained a model on rendered
images of six scene classes, carried all its weights over to real photographs, and compared the
result with the same model trained on the photographs alone. Their model was an ImageNet VGG16
convolutional base (14.7M parameters~\cite{simonyan2015}) with one dense layer, trained on a desktop
GPU with no quantisation. The ImageNet base is in both arms, so what they compared is ImageNet
initialisation with and without a synthetic stage. Their headline gain is \SI{+7.15}{\pp}; over the
working range their table gives about \SI{+1.6}{\pp}, which I use as the planning value
(\cref{sec:rel-bird}).

There are two reasons to expect a gain of that size to disappear at microcontroller scale. One is
capacity: for very small CNNs, Muralidharan et al.~\cite{muralidharan2026} find the difference
between pre-trained and from-scratch training usually falls inside run-to-run variance. The other is
precision: full-integer post-training quantisation~\cite{jacob2018,krishnamoorthi2018} costs
accuracy, and nothing guarantees that it costs two differently initialised models the same; here the
cost was of the same order as the gain in question. Neither says what the model costs to run, which
has to be measured on the device, or whether the device computes what the host scored.

This paper asks whether a synthetic pre-training stage, added on top of ImageNet initialisation,
changes the accuracy of a classifier small enough for a microcontroller, in float32 and in INT8;
what deploying it costs on an ESP32-S3 and an RP2040; and whether each board computes what the host
scored. The confirmatory tests and their thresholds were fixed before any data were collected
(\cref{sec:m-design}).

\paragraph{Arms and rows} Two labels run through the paper. An \emph{arm} is how a model is
initialised before it trains on the real photographs. Arm~A starts from random weights and arm~B
from ImageNet weights. Arms~C and~D start as A and~B do but train on the rendered images first. B
against D therefore isolates the synthetic stage on a pre-trained start, and it is the comparison
every registered test is built on. A \emph{row} is one model on a five-step size ladder
(\cref{tab:models}). Row~0 is a full-width MobileNetV2~\cite{sandler2018} at \SI{224}{px}, the
anchor. Rows 1 to~3 keep a \SI{96}{px} input and narrow MobileNetV2 from width 1.0 to 0.35. Row~4 is
a 61K-parameter CNN at \SI{48}{px}. No ImageNet weights exist for the CNN, so at row~4 arms B and~D
start from Imagenette weights and are written B$'$ and D$'$. Row~3 is the main deployable model and
carries the primary test, and row~4 carries the secondary. \Cref{sec:m-arms} gives the full
definitions.

\paragraph{Contributions} The paper makes four.
\begin{enumerate}
  \item \emph{The paired four-arm comparison in float32 and INT8.} Four initialisation arms,
  including from-scratch and ImageNet controls, are carried through post-training quantisation,
  paired by fold and replicated over seeds at the rows that carry the confirmatory tests. A
  difference-in-differences puts the quantisation-interaction question into a single number with
  a confidence interval.
  \item \emph{The sub-1M-parameter question, set against run-to-run variance.} For the smallest
  model the synthetic-stage effect is set against this pipeline's own run-to-run variance, the
  yardstick behind Muralidharan et al.'s claim.
  \item \emph{The on-device cost characterisation.} Latency, memory and energy per inference are
  measured across the ladder on two MCUs, including the ESP32-S3 arena's placement in internal
  SRAM or external PSRAM, measured both ways on the same model, and the RP2040's flash-capacity
  limit.
  \item \emph{Checking what the boards compute.} No hardware figure was used until the deployed
  model's INT8 outputs had been compared, element by element and with no tolerance, with
  TensorFlow Lite's reference kernels. The ESP32-S3 matched exactly once I had patched an
  out-of-bounds write in ESP-NN 1.3.1's depthwise kernel. The RP2040 departs from the reference
  at the global average pool, and the host interpreter that scored every accuracy does not
  reproduce the reference arithmetic either. Extending these checks, an exploratory addendum
  found three further ESP-NN defects, all reported to Espressif (\cref{sec:h-agree,sec:h-tool}).
\end{enumerate}

\Cref{sec:related,sec:methods} give the background and design,
\cref{sec:results,sec:fraction,sec:explore} the accuracy results, \cref{sec:hardware,sec:invariance}
the boards, and \cref{sec:discussion,sec:limits,sec:conclusion} close.

\section{Related work}\label{sec:related}
\subsection{Synthetic pre-training and the source study}\label{sec:rel-bird} Rendered and generated
images have stood in for labelled photographs for about a decade. Richter et al.~\cite{richter2016}
labelled frames from a commercial video game for segmentation, Kataoka et al.~\cite{kataoka2021}
pre-trained on formula-driven fractals and partly matched ImageNet pre-training after fine-tuning,
and Borkman et al.~\cite{borkman2021} trained a detector for 63 grocery products on 400,000
Unity-rendered images and then fine-tuned it on real photographs, against a detector trained on the
real photographs alone, the same synthetic-then-real against real-alone shape tested here. None of
them uses a model small enough for a microcontroller, and none quantises it to integers.

Bird et al.~\cite{bird2020} supply the dataset: six scene classes, with rendered images of virtual
environments and real photographs from the web. On an ImageNet VGG16 base they placed a single dense
layer at twelve widths, from 2 to 4096 neurons, and compared training on the rendered images and
then the photographs with training on the photographs alone. ImageNet is in both arms, so in the
notation here this is arm~B against arm~D; arms A and~C have no counterpart (\cref{sec:m-arms}).
They report an average final-accuracy gain of \SI{+7.15}{\pp}. That figure is harder to read than it
looks. The two arms were fine-tuned for different lengths, 50 epochs without the synthetic stage and
10 with it, and the average is carried by the ends of the width range: at 2 and 4 neurons the
non-transfer arm reaches only 18.69\% and 27.32\%, and from 1024 neurons up it stays below 70\%.
Over the six widths from 8 to 256 neurons, the working range set in the analysis plan before any
data were collected, the mean difference recomputed from their Table~II is \SI{+1.587}{\pp}, and
\SI{+1.613}{\pp} with the 512-neuron row added; both round to \SI{+1.6}{\pp} (Supplementary
Table~S3). They report no spread across folds, no duplicate screening of the photographs, no code
and no measure of how varied the rendered set is, and I found no third-party results on the dataset.

From the source study I take the dataset, the B-against-D comparison and the \SI{+1.6}{\pp} planning
value. The primary test asks whether the INT8 difference at row~3 excludes it. The anchor asks
whether a gain appears at all on a full-width MobileNetV2 at \SI{224}{px} with a quarter of the real
data; it is not a replication of their VGG16 model. The design deals with the other points directly:
every arm gets the same real-data budget, starting accuracy is reported only for arms that have a
synthetic stage, and every contrast carries its spread across folds and seeds.

\subsection{Transfer at small model scale}\label{sec:rel-scale}
Muralidharan et al.~\cite{muralidharan2026} scaled EfficientNet and MobileNetV3 backbones from a few
million parameters to under 20K and compared ImageNet pre-training with training from scratch at
each size. Below roughly 100K to 220K parameters the advantage was no longer reliably positive, and
for the smallest backbones it sat inside run-to-run variance. Their contrast is arm~B against arm~A;
the confirmatory question here is arm~D against arm~B, a synthetic stage on top of an initialisation
the model already has, and their result does not predict it either way. It does give a size scale.
MobileNetV2's last 1280-channel convolution is not width-scaled~\cite{sandler2018,lin2020}, so row~3
still has 0.42M parameters, above their threshold, while row~4, at about 61K, is below it and starts
from Imagenette rather than ImageNet. Bouthillier et al.~\cite{bouthillier2021} trace how much
benchmark variance comes from data splits and seeds, which supports pairing arms on the same split
and treating the split as the unit of replication, as here (\cref{sec:m-stats}).

Pre-training also speeds training: He et al.~\cite{he2019} matched ImageNet-pre-trained detectors
from random initialisation only with longer schedules, and Kornblith et al.~\cite{kornblith2019}
report faster fine-tuning from ImageNet weights. A fixed budget therefore favours the pre-trained
arms, and contrasts with the from-scratch arm~A are reported only descriptively. When a source stage
leaves a model worse off than the same model trained without it, the literature calls it negative
transfer~\cite{pan2010,zhang2023}; I use the term only where a synthetic stage is measured against
an already pre-trained baseline, at row~4 and the anchor. Finally, Catalan-Tatjer et
al.~\cite{catalantatjer2026} found in large language models that training history changes how much
accuracy post-training quantisation removes. Their models and bit widths are far from an 8-bit CNN,
but the premise motivates the difference-in-differences: two arms that differ only in training
history need not pay the same INT8 cost.

\subsection{On-device measurement}\label{sec:rel-device}
The runtime here is TensorFlow Lite Micro~\cite{david2021}. The application hands it one block of
memory, the tensor arena, which a planner lays out before the first inference, and each operator
runs through either a portable reference kernel or one the chip vendor has optimised. Lin et
al.~\cite{lin2021mcunetv2} found that executing the first blocks patch by patch cut peak memory
four- to eightfold: the arena a model needs depends on how the runtime executes it as well as on the
architecture; every latency here therefore carries its build, and arena placement is measured
(\cref{sec:m-hw}). Banbury et al.~\cite{banbury2021micronets} found the power of 400 small image
classifiers on Cortex-M boards almost constant from model to model ($\sigma/\mu = 0.00731$), so two
arms with the same latency should use about the same energy; I measured that instead of assuming it
(\cref{sec:invariance}).

Wilson's TinySSL~\cite{wilson2026} reports \SI{26.1}{\second} per inference for a width-0.35
MobileNetV2 in INT8 on an ESP32-S3 at \SI{128}{px}, without naming its kernels. My latency of record
for the same width at \SI{96}{px} is \SI{152.7}{\milli\second} (\path{main_T24_final_PRE_ARENA_PROFILE.cc}), 171 times lower, or about 96 times
after scaling for input area. Kernel choice could explain much of that: published figures on this
chip differ by a factor of 3.5 between two runtimes~\cite{faggi2026}, by 18 to 43 with and without
ESP-NN~\cite{espnn2026}, and by 49 to 61 from a naive C baseline to SIMD code~\cite{zhang2025}. I
could not obtain Wilson's model or build, and cannot explain the gap. Wilson also states that
hardware metrics do not depend on how a model was trained, which is checked on the device here
(\cref{sec:invariance}). Latency for small INT8 vision models on this chip has been published
before~\cite{espnn2026,wilson2026,faggi2026,zhang2025}, and of those only Zhang and Poonawala give
an energy figure, without saying for which model or how it was measured, so no priority is claimed
for the hardware results.

An optimised kernel is meant to compute what the reference kernel computes. TensorFlow Lite Micro's
documentation says most of its kernel tests require an optimised kernel to be bit-exact with the
quantised reference~\cite{tflmoptkernels}, but those tests run in the library's build, not on a
given board with a given model, and Schl{\"o}gl et al.~\cite{schlogl2023} found a fixed model and
input giving numerically different outputs across 75 CPU and GPU platforms. Here every deployed
model was compared with the reference kernels before its hardware figures were used
(\cref{sec:m-verify}).

\section{Methods}\label{sec:methods}
\subsection{Study design, registration and addendum}\label{sec:m-design}

\paragraph{Registration} The analysis plan was registered on the Open Science Framework (OSF) on 13
August 2026~\cite{yadav2026osf}. No training run came before that date; the earliest run in the
results file is dated 24 August 2026. I keep the split between confirmatory and exploratory analysis
that registration is meant to protect~\cite{nosek2018}, and report departures from the plan
following Willroth and Atherton~\cite{willroth2024}. The registration makes nine declarations: four
tests (\cref{tab:tests}), namely the primary at row~3, the secondary at row~4 (arms B$'$ and D$'$),
a difference-in-differences at row~3 and a three-way anchor criterion at row~0, and five rules,
covering a Holm correction over the primary and secondary only, the real-data epoch budget, adding
seeds, failed runs, and a window for amendment. The decision thresholds, including the
\SI{+1.6}{\pp} planning value (\cref{sec:rel-bird}), are part of the registered text
(\cref{sec:m-stats}). The results schema and the fold assignment were fixed before any grid run, and
the plan was never amended.

\paragraph{Registered rules applied} The epoch-budget rule gave 42 epochs (\cref{sec:m-train}). A
paired pilot gave $\sigma_d = \SI{1.14}{\pp}$, under the \SI{2}{\pp} that would have raised the
cells from three seeds to five. No run was excluded on its accuracy or any other outcome. One pilot
row was withdrawn before the grid opened because its value could not be reproduced from its retained
checkpoint, and the grid ran that position afresh. Each of the 430 declared cells has exactly one
row in the results file.

\paragraph{Departures} Every departure after the first main-grid run went into my project log as a
dated, numbered entry. Supplementary Table~S5 lists those that bear on a reported result, with the
corrections made to values first filed there. Six matter for reading the paper. The first
epoch-budget pilot ran into a batch-normalisation fault and was discarded, and the pilot was re-run
on a corrected recipe before any grid run (\cref{sec:m-train}). The registration calls the
data-fraction sweep's figure confirmatory but registers no test on it, so I report the sweep
descriptively. For a failed anchor the registration says the paper becomes ``the characterisation
plus the bounded absence''~\cite{yadav2026osf}; my headline also reports how the effect changes with
the share of real data (\cref{sec:fraction}), with the registered results first and unchanged.
Energy on the RP2040 was brought into scope after registration (\cref{sec:m-hw}). Two addendum
items were not run and one check was not done, as set out below. And the repeat captures on both
boards (items ESP7 and~P6) missed two of their declared conditions. The addendum asked for the
supply voltage to be measured, which I did not do (\cref{sec:m-hw}), and it named the ladder harness
for the ESP32-S3 repeat latencies, which ran on a harness of their own (\cref{sec:h-esp}).

\paragraph{Addendum} After the registered analysis had been read, I filed an exploratory addendum on
OSF on 5 October 2026 (\url{https://doi.org/10.17605/OSF.IO/CB7X3})~\cite{yadav2026osfaddendum}. It
is a new registration that names the original, which was not updated, and it was filed before any of
its items ran. It changes none of the nine declarations. It declares four groups of work: more
replicates for cells the registration ran with one seed or one synthetic draw (the anchor, the
data-fraction sweep and the trend regression); controls on the row-4 gain with scarce real data, a
second training recipe and a determinism check; re-scoring of every registered INT8 model on both
kernel paths, with analyses and a data-hygiene screen that needed no new runs; and repeat and
extended hardware measurements on both boards. I cite each item by its code. C2 to C12, C14 and C15 are training and host items (there is no
C13, which was dropped before filing, and C1, C16 and C17 were setup, desk and presentation work
with no result of their own), ESP1 to ESP12 are the ESP32-S3 items (filed as E1 to E12) and P1 to
P7 are the RP2040 items. Supplementary Table~S6 gives each item's declaration, seeds, count, trigger and
status. New runs used hash-checked copies of the registered code, with the same recipe, folds and
42-epoch budget unless an item varied one on purpose, and a results file of their own. Their seeds
run from 921 to 940, and arms B and~D share a seed wherever an item ran both in one cell. C6 is the
exception by design, since it repeats seed~1, and in C7 the two new synthetic draws (seeds 921
and~922) and the two extra arm-B runs (923 and~924) are separate replicates. Where an item reused registered seed-1 runs, those
rows were read unchanged.

Two items were not run. The row-0 arm of the shuffled-label control (C10) was to run only if the
four-draw anchor interval lay wholly below zero; its upper end was \SI{+0.031}{\pp}, so the arm was
not run. The RP2040 rebuild with reference kernels in place of CMSIS-NN (P3) was not run, because
the RP2040's first departure from the reference is at the global average pool (MEAN), which is not a
CMSIS-NN kernel. The meter check on the RP2040 against a reference resistor, the second part of item 
P1, was not done. Every addendum result is tagged 
\rc{exploratory} and placed after the registered result it bears on. The registered value comes 
first and stays of record, tables keep the two in separate rows or columns, and where they disagree 
I report both. Where an item left a reading open, the text says so beside the number.

\paragraph{Result classes and families} The four registered tests are \emph{registered
confirmatory}; for the anchor the registered criterion is float32; INT8 is measured and exploratory.
The trend regression is \emph{registered optional} and the data-fraction sweep a \emph{descriptive
registered measurement}; where the addendum added replicates to a seed-1 cell, the seed-1 value
stays of record. Everything else is \emph{exploratory}, apart from RP2040 energy, a
\emph{post-registration addition}, and one paragraph of the Discussion, a \emph{hypothesis}.

Three statistical families are kept apart. The registered Holm family holds the primary and
secondary tests and nothing else, at a family-wise level of 0.05. The registered
difference-in-differences is a family of its own. The 16 accuracy contrasts declared in the addendum
form a Benjamini--Hochberg screen at a false discovery rate of 0.05, two-sided, using the corrected
$t$-test of \cref{sec:m-stats}; \cref{sec:explore} gives every one with its raw and adjusted $p$-values.
The screen points to differences that deserve a direct test and is not a test itself. No addendum
result enters a registered family. The hardware measurements were not registered. The agreement
checks, the arm-invariance test and the RP2040 boundary test each had a pass criterion fixed before
the device runs, and everything else on the boards is reported as a measurement.

\paragraph{Run counts} The final design budget is 476 runs, 472 of them on the 42-epoch real-data
budget, and took 18.89 GPU-hours. Its 430 declared cells are all complete. The results file holds
547 rows; 73 of the 117 outside the declared cells are exploratory follow-ups, also outside the
budget. The addendum added 710 training runs, all completed and none excluded, in a results file of
their own and outside every registered count. Supplementary Table~S5 breaks the budget down.

\subsection{Dataset}\label{sec:m-data}
All images come from the public Kaggle release of Bird et al.'s dataset~\cite{bird2020data}. It
holds 3,013 real photographs (JPEG, \SI{128}{px}) and 5,928 rendered images (PNG, \SI{256}{px}) in
six classes. The paper describes 6,000 rendered images and balanced photographs~\cite{bird2020}; in
the release, field and living room have only about 60\% as many photographs as bathroom. My class
indices are alphabetical, from bathroom (0) to staircase (5), and each row resizes both kinds of
image to its own input size (\cref{sec:m-train}). Supplementary Note~A gives the counts by class.

The photographs came from the web and Bird et al.\ report no duplicate screening, so I screened them
before building the folds: an MD5 hash for exact copies, and an $8 \times 8$ perceptual
hash~\cite{imagehash} with a threshold of 5 bits for near-copies. The screen removed 18 photographs
(0.60\%), 14 exact and 4 near, all in pairs within one class, which leaves 2,995. Any threshold from
4 to 9 bits removes the same four near-copies (Supplementary Note~A). A mirror image or a heavy crop
would get past the screen, and I could not check overlap with ImageNet (\cref{sec:limits}).

The rendered images were captured as camera sweeps of 288 frames, 5\textdegree{} apart at four
tilts~\cite{bird2020}. From the capture times in the file names I recovered 23 camera placements, 17
complete sweeps and 6 truncated ones, or about 258 images per placement; the four in computer lab
are inferred, not located (Supplementary Note~A). The rendered set is therefore less varied than its
size suggests.

\paragraph{Addendum: duplicate screen \rc{exploratory}} Repeating the screen at 8 and 12 bits (item
C15) found nothing at 9 bits or fewer. At 12 bits it found 18 pairs, each split across two folds and
ten of them across two classes, which I did not inspect by eye. Dropping both members of every pair
from their test folds moved neither INT8 confirmatory contrast by more than \SI{0.03}{\pp}. Float32
test predictions were never saved, so the anchor and the difference-in-differences could not be
rechecked this way.

The folds apply to the photographs only. The rendered images enter training only through the
synthetic stage, whose checkpoint is shared by all ten folds and every data fraction
(\cref{sec:m-arms}), so no test photograph reaches training that way. I sorted the 2,995 photographs
by file path and split them with scikit-learn's \texttt{StratifiedKFold} (version 1.6.1, shuffled,
\texttt{random\_state} 42)~\cite{pedregosa2011}. Folds 0 to~4 hold 300 photographs and folds 5 to~9
hold 299. Each run tests on one fold and trains on the other nine, with no validation split. One
image is worth about \SI{0.33}{\pp}. The folds are stratified by class and not grouped by scene; a
spot check of 30 photographs from two classes found no repeated scenes. The fold assignment is
recorded with its hash and is available on request.

\subsection{Model ladder and arms}\label{sec:m-arms}
The ladder has five rows (\cref{tab:models}). Rows 0 to~3 are MobileNetV2~\cite{sandler2018} and
row~4 is a small CNN. Row~0, at full width and \SI{224}{px}, is the anchor. It checks whether a gain
of the kind Bird et al.\ report appears in this pipeline at all, and it was not deployed in the
registered design (the addendum later ran it on the ESP32-S3; \cref{sec:h-esp}). Rows 1 to~3 keep
the input at \SI{96}{px} and narrow the network from width 1.0 to 0.35. Row~3 carries the primary
test and is the main deployable model. Row~4 takes a \SI{48}{px} input through four $3 \times 3$
convolutions with 16, 32, 64 and 64 filters, ReLU and same padding, with $2 \times 2$ max-pooling
after the first three and no batch normalisation. Every row ends in global average pooling and a
six-way dense layer with softmax and no dropout.

\begin{table}[htbp]
\centering
\caption{The model ladder. Rows 3 and 4 carry the confirmatory tests and row~0 the
anchor.}
\label{tab:models}
\small
\begin{tabular}{@{}c l r r r l c@{}}
\toprule
Row & Model & \makecell[r]{Input\\(px)} & Parameters & \makecell[r]{MACs\\($10^6$)} &
\makecell[l]{Start for\\arms B and D} & \makecell{Seeds per cell\\(A, C / B, D)} \\
\midrule
0 & MobileNetV2, width 1.0 & 224 & 2,265,670 & 299.50 & ImageNet & 1 / 1 \\
1 & MobileNetV2, width 1.0 & 96 & 2,265,670 & 55.02 & ImageNet & 1 / 1 \\
2 & MobileNetV2, width 0.5 & 96 & 713,910 & 17.61 & ImageNet & 1 / 1 \\
3 & MobileNetV2, width 0.35 & 96 & 417,894 & 10.66 & ImageNet & 1 / 3 \\
4 & CNN (this study) & 48 & 60,902 & 7.63 & Imagenette & 1 / 3 \\
\bottomrule
\end{tabular}
\par\smallskip\raggedright\footnotesize\noindent\textit{Notes:} Parameters are for the
whole model, six-way head included. MACs are the multiply--accumulate operations of
the convolution and dense layers in one inference, counted from the layer shapes.
At row~4 arms B and D start from Imagenette weights and are written B$'$ and D$'$.
Rows 1 to~4 were built for both boards, though row~1 does not fit in the RP2040's flash;
row~0 was built only in the addendum, for the ESP32-S3.
\end{table}

Arm~A starts from random weights and arm~B from ImageNet weights, and both then train on the
photographs alone. Arms~C and~D start as A and B do, but train on the 5,928 rendered images first.
The ImageNet weights are the Keras MobileNetV2 files~\cite{chollet2015keras}, trained on
ILSVRC-2012~\cite{russakovsky2015}, one per width and input size, checked against recorded hashes.
\Cref{fig:arms} shows the final accuracy of all four arms at every row. B against D tests the
synthetic stage and is the comparison Bird et al.\ made (\cref{sec:rel-bird}). At full data arm~D
takes about 10\% more optimiser steps than arm~B (372 synthetic steps on top of 3,570), which I
accept as part of the treatment. The anchor trains B and D at row~0 on 25\% of the photographs, and
the data-fraction sweep trains B and D at rows 3 and~4 on 10, 25 and 50\%, all with one seed.

No ImageNet weights exist for the CNN, so at row~4 arms B and D start from weights I pre-trained on
Imagenette~\cite{howard2019imagenette}, a ten-class subset of ImageNet with 9,469 training images,
for 40 epochs with flips as the only augmentation. I write these arms B$'$ and D$'$. Within row~4
they are paired as cleanly as B and D elsewhere. Across rows the corpus and the model shrink
together. Before either pre-training ran I set a floor of 30\% top-1 accuracy on Imagenette's
validation split, against 10\% for chance; the CNN reached 61.2\% and the MobileNetV2 used in the 
bridge below reached 51.8\%. To size the cost of the weaker start, a calibration bridge pre-trains the row-3 MobileNetV2
on Imagenette in the same way and runs it as arm~B at row~3 over the ten folds with one seed
(\texttt{init\_source='imagenette'}). That MobileNetV2 overfits far more than the CNN, with a gap of
17.3 points between training and validation accuracy against 1.5, and it sees \SI{96}{px} inputs
where the CNN sees \SI{48}{px}. So the bridge bounds the downgrade as it actually happened, not the
effect of corpus size alone. Its result is with the secondary test (\cref{sec:r-conf}).

\subsection{Training recipe}\label{sec:m-train}
Every run used one training recipe, the same for every arm and row, hashed and frozen before the
first grid run. Grid training ran on Google Colab on an NVIDIA T4 GPU with TensorFlow
2.20.0~\cite{tensorflow2015} and Keras 3.13.2~\cite{chollet2015keras}. The optimiser is
Adam~\cite{kingma2015} at a constant learning rate of $10^{-4}$ with the Keras defaults for its
other settings, no weight decay and no schedule; the batch size is 32 and the loss sparse
categorical cross-entropy. The rate is the same in every arm, so no arm gets a tuned setting the
other lacks. It is $10^{-4}$ rather than $10^{-3}$ so that end-to-end fine-tuning keeps more of the
ImageNet start, and it is constant so that the budget rule, which compares five-epoch averages, sees
no scheduled step. The whole network trains in the real-data stage, in every arm.

Each image is resized bilinearly to the row's input size, without anti-aliasing, and scaled to $[-1,
1]$ as $x/127.5 - 1$. The only augmentation is a random horizontal flip, and the training set is
reshuffled every epoch. A run's seed fixes the initialisation of the new head (and of the whole
network in arms A and~C), the shuffle order and the flips. In arms C and~D the real-data stage is
re-seeded when it starts. TensorFlow's deterministic-ops mode was off, so two runs with the same
seed can differ a little, and the spread between runs in \cref{sec:r-seed} is run-to-run variance,
not seed variance alone. For the anchor and the sweep the nine training folds are cut down class by
class, from a fixed seed and the fold index, so arms B and D at the same fold and fraction train on
the same photographs.

Keras builds MobileNetV2's batch-normalisation layers with a momentum of 0.999. With 85 steps per
epoch the moving statistics then trail the weights by many epochs, and a randomly initialised
network evaluated in inference mode put every test image in one class. I use 0.9 in every
batch-normalisation layer of every arm. Left in place, the fault would have favoured the pre-trained
arms (Supplementary Note~B).

The synthetic stage uses the same settings on all 5,928 rendered images, with no validation split,
for two epochs of 186 batches. Its six-way head carries over into the real-data stage; the optimiser
state does not. At the hand-over, arms C and~D are scored on the test fold in inference mode, and
that is the starting accuracy I report. Arms A and~B have an untrained head at that point, so they
get no starting accuracy, which avoids the artefact described in \cref{sec:rel-bird}.

The length of the synthetic stage and whether its base is frozen were left open by the plan and
fixed before the grid, on row~3, arm~D, folds 0 to~2, with their own seeds (701 to 703) and decision
rules on float32 accuracy written before the runs. Both experiments ended in ties under their rules,
so the cheaper setting held: two synthetic epochs, against 5, 10 and 50, and a trainable base, with
frozen minus unfrozen at \SI{+0.22}{\pp} (95\% CI \num{-0.65} to \num{+1.09}, an ordinary $t$
interval over nine fold--seed pairs). Arm~C uses the same
settings. Supplementary Note~B and Supplementary Figure~S2 give both experiments.

The real-data stage lasts 42 epochs in every arm and row. The registered rule finds the first epoch
$E \geq 10$ at which the mean test accuracy over the last five epochs beats the five before by less
than 0.5 points, and sets the budget at $E + 10$~\cite{yadav2026osf}. Applied to a 100-epoch pilot
of arm~A on row~3, fold~0, it gave $E = 32$; a check at row~4 gave $E = 31$. The accuracy reported for a run is its
test-fold accuracy after the last epoch, never the best epoch; there is no early stopping. Accuracy
after epoch~5 and macro-averaged F1 are stored and reported descriptively. Every test, threshold and
planning value is in accuracy points.

\subsection{Quantisation}\label{sec:m-quant}
Every run is scored twice on its test fold, as the trained float32 model and as an INT8 copy made
straight after training from the last-epoch weights. The copy comes from full-integer post-training
quantisation with the TensorFlow 2.20.0 converter~\cite{tensorflow2015}, with the default
optimisation flag, a representative set (below), the operator set restricted to INT8 built-ins and
INT8 input and output, so the converter stops rather than leave an operator in floating
point~\cite{litertptq}. All five rows go through the same conversion. Read from the four deployed
files, the weights are INT8 with one scale per output channel and a zero point of
0~\cite{krishnamoorthi2018}, activations have one scale and zero point per tensor, and biases are
32-bit integers~\cite{jacob2018}. The input scale is $2/255$ with a zero point of $-1$, and the
output scale $1/256$ with a zero point of $-128$. Calibration is the representative set's activation
ranges and nothing more: none of the cross-layer equalisation, bias correction or AdaRound steps of
Nagel et al.~\cite{nagel2021}, and no quantisation-aware training.

Each fold has one representative set of 200 photographs, stratified by class and drawn from its nine
training folds with a seed fixed by the fold index. Within a fold it is the same set for every arm,
row, seed and data fraction. Holding it fixed across arms keeps calibration noise out of the paired
B-against-D difference. Drawing it from the full training split at every fraction means the sweep
varies the training data alone. A rule fixed in advance chose the size and stratification on
stability across six candidates, which all came within 0.6 images of each other (Supplementary
Note~C). Which weights were quantised mattered far more: five arm-B runs at row~3, fold~0, lost 9,
$-1$, 7, 0 and 5 images to quantisation (mean \SI{1.33}{\pp}, standard deviation \SI{1.45}{\pp}).

All 545 INT8 accuracies in the results file come from this policy. Five were recorded before it was
fixed and were recomputed from retained weights; the two plateau pilots have none. The conversion is
not byte-reproducible. Thirteen conversions of the same weights gave thirteen MD5 hashes, and some
lengths differed by 8 bytes of tensor names, while the INT8 accuracy matched every time. That
variation is already inside the recorded accuracies and inside $\sigma_d$. The deployed files were
rebuilt from retained weights, are identified by hash, and each scored exactly the value recorded
for its run.

INT8 accuracy was scored with TensorFlow 2.20.0's \texttt{tf.lite.Interpreter} at default settings,
which I call the default path. It hands most of the network to one delegate and does not reproduce
TensorFlow Lite's reference kernels, the reference path, which the ESP32-S3 matches exactly. On the
deployed row-3 file the reference path scored 265 of 300 images against 262 on the default path
(\SI{+1.0}{\pp}); six predictions changed, three of them on ties, and 65 of the 300 output vectors
differed by more than one quantisation step. Every grid run used the default path, so its offset is
inside the recorded accuracies and $\sigma_d$, and it largely cancels in a paired difference.

\paragraph{Addendum: both paths for every model \rc{exploratory}} A host scorer on the reference
kernels reproduced all 120 device output values on the twenty row-3 images of \cref{sec:h-agree}
(item C2). All 430 registered INT8 models were then rebuilt from their retained weights and
re-scored on both paths (item C5), and every rebuilt file gave its recorded default-path count.
Reference minus default averaged \SI{-0.009}{\pp}, with a standard deviation of \SI{0.753}{\pp} and
a range of $-13$ to $+8$ images; the two paths disagreed most for arm~A and at row~2. On the
reference path the primary became \SI{+0.245}{\pp} (95\% CI \num{-0.873} to \num{+1.363}), the
secondary \SI{-0.601}{\pp} (\num{-0.906} to \num{-0.296}) and the difference-in-differences
\SI{-0.200}{\pp} (\num{-1.492} to \num{+1.092}). None moved by more than \SI{0.16}{\pp}, and none
reads differently. These are sensitivity re-estimates outside every family, and the default-path
values stay of record.

\subsection{Statistical analysis}\label{sec:m-stats}
Every test is paired over the ten folds. Within a fold the contrast is the difference in last-epoch
accuracy, in percentage points, $\Delta = D - B$ (row~4: $D' - B'$); a positive value means the
synthetic stage helped. Three-seed cells are averaged within the fold before the difference is
taken. Accuracies are handled as counts of correct images. I write $\bar\Delta$ and $s$ for the mean
and standard deviation of the ten differences, and ``folds above zero'' counts those strictly above
zero. \Cref{tab:tests} lists the tests.

\begin{table}[htbp]
\centering
\caption{The registered tests and the registered optional trend regression.}
\label{tab:tests}
\small
\begin{tabular}{@{}>{\raggedright\arraybackslash}p{1.9cm}
                  >{\raggedright\arraybackslash}p{3.3cm}
                  >{\raggedright\arraybackslash}p{2.6cm}
                  >{\raggedright\arraybackslash}p{4.0cm}
                  >{\raggedright\arraybackslash}p{2.0cm}@{}}
\toprule
Test & Contrast & Data & Decision rule & Multiplicity \\
\midrule
Anchor & Row~0, float32, B against D & 25\% of real data, one seed &
  Pass if the lower bound is above 0; fail if the upper bound is below $+1$; otherwise
  inconclusive & Outside Holm \\
Primary & Row~3, INT8, B against D & Full data, three-seed means &
  95\% interval; a bound if it excludes $+1.6$ & Holm \\
Secondary & Row~4, INT8, B$'$ against D$'$ & Full data, three-seed means &
  Corrected $t$-test & Holm \\
Difference-in-differences & Row~3, float32 minus INT8 & Full data, three-seed means &
  95\% interval; ``quantisation removes at most $X$ points of the transfer benefit'' &
  Own family \\
Trend regression & Slope of $\Delta$ on $\log_{10}$ parameters & Full data, seed~1 &
  Corrected test on ten slopes & Outside Holm; optional \\
\bottomrule
\end{tabular}
\par\smallskip\raggedright\footnotesize\noindent\textit{Notes:} Differences are in
percentage points, and every interval is a two-sided 95\% interval. Row~0's registered
criterion is float32; its INT8 values are measured and exploratory (\cref{sec:explore}).
The difference-in-differences at row~4 is descriptive only.
\end{table}

The registration fixed the order of reporting: the effect with its confidence interval, then a
corrected $t$-test, then a rank test as a robustness check~\cite{yadav2026osf}. In ten-fold
cross-validation any two training sets share eight of their nine folds, about 89\% of
each~\cite{dietterich1998}, so an
ordinary paired $t$-test on the folds is too liberal. I use the corrected resampled $t$-test of
Nadeau and Bengio~\cite{nadeau2003}, which widens the standard error to
\begin{equation}\label{eq:nb}
  \mathrm{SE} = s\,\sqrt{1/10 + 1/9},
\end{equation}
1.453 times the uncorrected $s/\sqrt{10}$. Intervals are
$\bar\Delta \pm 2.262\,\mathrm{SE}$, and $p$-values are two-sided with 9 degrees of freedom. The
rank
test is an exact Wilcoxon signed-rank test with average ranks for ties and zero differences
dropped, reporting the smaller rank sum $W$. It makes no allowance for the
overlap~\cite{demsar2006}, so where the two disagree I go with the corrected $t$-test.

Holm's procedure~\cite{holm1979} covers the primary and secondary tests only. For the primary, an
INT8 interval at row~3 that excludes the $+1.6$ planning value is itself the finding, an
equivalence-style bound and not a claim that the difference is zero.

The difference-in-differences is formed fold by fold from the float32 and INT8 accuracies of the
same trained models,
\begin{equation}\label{eq:did}
  \Delta_{\mathrm{DiD}} = \Delta_{\mathrm{f32}} - \Delta_{\mathrm{INT8}},
\end{equation}
which is arm~D's quantisation cost minus arm~B's. The registered reading, ``quantisation removes
at most $X$ points of the transfer benefit''~\cite{yadav2026osf}, assumes a float32 benefit to
remove; at row~4 the registration reports the contrast descriptively only, and as interpretable
only if that benefit is non-zero. Where there is no float32 benefit I report the interval as a bound on how far
quantisation moves the difference.

The registered pilot estimated $\sigma_d$ at \SI{1.14}{\pp} directly from five seeds of both arms on
row~3, fold~0. The registration's fallback relation, $\sigma_d \approx
\sqrt{2}\,\sigma_{\mathrm{seed}}$, assumes uncorrelated arms (in the pilot $r = 0.63$) and would
have given 1.80 to \SI{2.07}{\pp}. To set effects against run-to-run variance, the yardstick
Muralidharan et al.\ use~\cite{muralidharan2026}, I also report a pooled run-to-run standard
deviation: for each arm the root mean square over folds of the standard deviation across seeds, then
the two arms pooled the same way. On the grid, $\sigma_{\mathrm{DiD}}$ came out at \SI{1.238}{\pp}
against \SI{1.116}{\pp} for $\sigma_d$ in the primary, so the difference-in-differences is not the
sharper measure. The arms' float32 cell means correlate at $r = 0.954$ across folds and their INT8
cell means at only 0.675, so the cost of quantisation is weakly shared between arms.

The registered optional regression fits, within each fold, an ordinary least-squares line of
$\Delta$ against $\log_{10}$ of the whole-model parameter count and summarises the ten slopes, in
points per decade, with the same corrected interval. Float32 uses rows 0 to~4 and INT8 rows 1 to~4,
as declared before the fit, and every row enters at seed~1. Rows 0 and~1 share a parameter count but
differ in input size and ImageNet file, and row~4 starts from Imagenette, so the slope does not
isolate capacity. It was run after the confirmatory results had been read.

The folds replicate the real-data stage and the test images, not the synthetic stage, since one
synthetic checkpoint serves all ten folds. Rows 3 and~4 have three draws of that stage through their
seeds and row~0 has one, which matters most for the anchor (\cref{sec:limits}). The $5 \times 2$
cross-validation named as a contingency in the registration~\cite{dietterich1998,yadav2026osf} was
not run.

\subsection{Hardware and energy measurement}\label{sec:m-hw}
For each of rows 1 to~4 I deployed one trained model: the arm-B run (B$'$ at row~4) on fold~0,
seed~1, with the full real data. Row~0 was not built for either board in the registered
measurements. For the arm-invariance test I also built the row-3 arm-A and arm-D models from the
same fold and seed (\cref{sec:invariance}). The measurements of record ran one input, a photograph
of a living room from fold~0's test split, compiled into the firmware; several addendum checks ran
twenty photographs instead.

\paragraph{ESP32-S3} The main board was an ESP32-S3-N16R8 with \SI{16}{\mega\byte} of
flash and \SI{8}{\mega\byte} of octal PSRAM at \SI{40}{\mega\hertz}, running at
\SI{240}{\mega\hertz}. The firmware used ESP-IDF 6.0.2~\cite{espidf602}, TensorFlow Lite
Micro~\cite{david2021} through esp-tflite-micro 1.4.0~\cite{esptflitemicro140}, and ESP-NN
1.3.1~\cite{espnn131} with a local patch to its tiled depthwise kernel (\cref{sec:h-tool}), which is
available on request. Each build reported ESP-NN's status in its boot log. The tensor arena was one
fixed block, a statically linked array in internal SRAM or a 16-byte-aligned heap block in PSRAM,
and each build read its location from the pointer at start-up. The largest static array that would
link was \SI{290019}{\byte}, so rows 1 and~2 ran from PSRAM and rows 3 and~4 from SRAM. Two controls
separate placement from size: row~4 rebuilt with its arena in PSRAM at the same reserved size, and
an earlier row-3 build that moved the arena to PSRAM and also raised its reserved size from
\SI{281600}{\byte} to \SI{352256}{\byte}.

\paragraph{Latency} Each latency run made one untimed warm-up inference and then 20 timed calls to
TFLM's \texttt{Invoke()}, bounded by two reads of the ESP-IDF microsecond timer. TFLM's default
memory planner, the online greedy one, reuses a tensor's memory once its last reader has
run~\cite{david2021}, and in esp-tflite-micro 1.4.0 the planner's header states that it does not
preserve all tensors after an invocation~\cite{esptflm140planner}. A project contributor confirmed
in 2023 that this covers the model input~\cite{tflmissue2290}. In the row-3 model only the first
operator reads the input, so the firmware copies it back before each call, outside the timed
region. MicroTFLite 1.0.4 on the RP2040 runs the same lifetime code. I report the median and the interquartile
range (Tukey hinges). Firmware edits that left the model alone moved the median by
\SI{48}{\micro\second} at row~3 and \SI{159}{\micro\second} at row~2, so a latency means something
only within its build. The ladder takes all four rows from one lineage (\texttt{main\_T36.cc}).
Row~3's latency of record comes from an earlier harness (\path{main_T24_final_PRE_ARENA_PROFILE.cc})
and is quoted on its own. Timing windows from the energy firmware are never used as latencies, and
each addendum harness is a lineage of its own.

\paragraph{RP2040} The second board was a Raspberry Pi Pico~H (RP2040, two Cortex-M0+ cores), used
as a boundary test. The firmware used the arduino-pico core 5.6.0~\cite{arduinopico560} (GCC 14.3)
and MicroTFLite 1.0.4~\cite{microtflite104} with its all-operators resolver; the kernels are
CMSIS-NN with no acceleration path and come from a different TFLM snapshot from the ESP32-S3's. The
clock of record, \SI{133}{\mega\hertz}, was enforced at compile time and checked at boot. Before any
build I fixed the possible outcomes for a row: flash overflow or RAM overflow at link time, failure
to allocate the arena, or a run. The linker gives \SI{262144}{\byte} of RAM, and the largest arena
that linked was \SI{244840}{\byte}. Rows 2, 3 and~4 were given \SI{235000}{\byte},
\SI{208896}{\byte} and \SI{131072}{\byte}, and the arena-used figures are planner readings, not
minima. Latency followed the ESP32-S3 protocol with \texttt{ModelRunInference()} and
\texttt{micros()}, and the library's floating-point conversion of inputs and outputs sits outside
the timed region.

\paragraph{Arm invariance} The row-3 arm-A and arm-D models went into the ladder firmware with
only the model-selection setting changed, and were compared with arm~B from the ladder. The
criteria were fixed before any device run: arenas identical in size and location, model files
the same size, latency medians within \SI{\pm 100}{\micro\second} of one another, and each arm's
outputs equal to its own reference, which checks that the right weights were linked. The
\SI{\pm 100}{\micro\second} bound comes from how far link layout moves latency, so a pass shows
invariance at that resolution only. Energy for the three arms was captured in one session, in
the order A, B, D, A, with
\begin{equation}\label{eq:armR}
  R = \frac{\bar{E}_{A1} - \bar{E}_{D}}{\bar{E}_{B}}
\end{equation}
on gross energy per inference, where $\bar{E}$ is a capture's mean over its 30 windows and $A1$
is the first arm-A capture. The test passed if $|R| \leq 1\%$, provided the half-width of its
95\% interval was at most 0.5\% and the two arm-A captures differed by at most 1\% of
$\bar{E}_{B}$; otherwise it was inconclusive. The 1\% bound is a judgement made before I had
measured any energy noise on this rig. $R$ on idle-subtracted energy is reported but does not
decide the test.

\paragraph{Energy rig} A Nordic Power Profiler Kit~II~\cite{ppk2} powered each board as a source
meter at \SI{3300}{\milli\volt} and recorded its current at \num{100000} samples per second. On both
boards the supply went into the 3V3 pin, downstream of any regulator, and the USB cable was
unplugged during capture. One GPIO line was held high for each inference and another for the whole
run of 30. On the ESP32-S3 a window holds one \texttt{Invoke()}, two timer reads and the two GPIO
writes that make its edges. After each window the firmware checked the output, restored the input
and idled for \SI{2}{\second}: a FreeRTOS delay on the ESP32-S3, with no power management and the
radios never started, and a wait-for-event with the clocks running on the RP2040. A capture was
accepted only with exactly one framing pulse and 30 inference pulses inside it. \Cref{tab:hwcond}
summarises the conditions of record. Supplementary Table~S4a lists every capture, with its file
hash where one is on record; the eight captures of the two failed RP2040 sessions have none. S4b
lists the sanity checks I ran after the figures were known, and S4c the build conditions and hashes.

\begin{table}[htbp]
\centering
\caption{Measurement conditions for the energy figures of record.}
\label{tab:hwcond}
\small
\begin{tabular}{@{}>{\raggedright\arraybackslash}p{2.6cm}
                  >{\raggedright\arraybackslash}p{5.9cm}
                  >{\raggedright\arraybackslash}p{5.3cm}@{}}
\toprule
 & ESP32-S3 & RP2040 (Pico H) \\
\midrule
Board & ESP32-S3-N16R8 board; \SI{16}{\mega\byte} flash,
  \SI{8}{\mega\byte} octal PSRAM & Raspberry Pi Pico~H \\
Clock & \SI{240}{\mega\hertz} & \SI{133}{\mega\hertz} \\
Runtime and kernels & ESP-IDF 6.0.2, esp-tflite-micro 1.4.0, ESP-NN 1.3.1 with a local
  patch & arduino-pico 5.6.0, MicroTFLite 1.0.4, CMSIS-NN without acceleration \\
Arena & Internal SRAM (rows 3, 4) or PSRAM (rows 1, 2), checked at boot &
  Internal SRAM \\
Supply & \SI{3300}{\milli\volt} set, not measured, into the 3V3 pin, after the board's regulator &
  \SI{3300}{\milli\volt} set, not measured, into the 3V3 pin; on-board regulator shut down \\
Window & One \texttt{Invoke()} & One \texttt{ModelRunInference()} \\
Between windows & Post-inference checks, then \SI{2}{\second} FreeRTOS delay; radios
  never started & Post-inference checks, then \SI{2}{\second} wait-for-event; no radio
  fitted \\
Sampling & \num{100000} samples/s; 30 windows per capture & Same \\
Warm-up & None; first window 5.9 to \SI{6.7}{\second} after power-on &
  \SI{1800}{\second} of inferences before capture \\
Drift guard & 0.225\% idle-subtracted (rows 1, 2, 4); 0.58\% gross (row~3 session) &
  $+0.107$\% gross, $+0.067$\% idle-subtracted \\
Status & Within the registered scope &
  Post-registration addition \\
\bottomrule
\end{tabular}
\par\smallskip\raggedright\footnotesize\noindent\textit{Notes:} The USB cable was unplugged during
capture on both boards. The full conditions are in Supplementary Table~S4c.
\end{table}

\paragraph{Energy per inference} For window $i$ the gross energy is
\begin{equation}\label{eq:egross}
  E_{\mathrm{gross},i} = V \sum_{k \in i} I_k \, \Delta t,
\end{equation}
with $\Delta t = \SI{10}{\micro\second}$ and $V = \SI{3.300}{\volt}$, the nominal set supply and
not a measured one, in every capture, addendum included. An error in it cancels in ratios between
captures but not in absolute millijoules. The idle current $\bar{I}_{\mathrm{idle}}$ is the mean over the 29 gaps between
windows, leaving out \SI{250}{\milli\second} at each end of every gap, longer than the work done
after an inference on either board. The idle-subtracted energy is
\begin{equation}\label{eq:einc}
  E_{\mathrm{inc},i} = \left(\bar{I}_i - \bar{I}_{\mathrm{idle}}\right) V \, t_i,
\end{equation}
where $\bar{I}_i$ and $t_i$ are the window's mean current and length. Each figure is the mean over
all 30 windows of one capture, from one fixed analysis script that a separate implementation
reproduced. The headline is idle-subtracted energy with gross energy beside it. Idle-subtracted
means above this firmware's own idle, which includes the power LED, PSRAM standby, the on-chip
USB block and the meter's logic inputs, none of which I separated. Every session of record had a
drift guard: the same firmware captured early and late had to agree within 1\%, a failed session could
be re-run once after its cause was written down, and the bound was never widened. ESP32-S3 row~3
comes from the arm-invariance session and rows 4, 2 and~1, with the row-4 PSRAM control, from a
second session that opened and closed on the row-3 firmware. The ESP32-S3 captures of record
started from a cold boot and the RP2040's followed \SI{1800}{\second} of warm-up. The addendum's
repeat captures had \SI{1800}{\second} of warm-up on both boards and one session per build, with
no drift guard, and were checked for dropped samples.

\paragraph{RP2040 energy} Energy on the RP2040 is a post-registration addition. The figures of
record supply \SI{3300}{\milli\volt} directly to the 3V3 pin with the 3V3\_EN pin tied to ground,
which turns the on-board regulator off. Raspberry Pi's documents rule this route
out~\cite{rpihwdesign,picodatasheet}, and I took it against their guidance, with the firmware
checking at boot and again before capture that VSYS was unpowered. It gives both boards the same
kind of supply, and its drift guard passed (\cref{tab:hwcond}). Two sessions supplied through VSYS
failed the drift guard, and a third with the regulator held in PWM mode passed but was dominated by
the regulator's own draw. Their figures are flagged in Supplementary Table~S4a and never pooled, and
Supplementary Note~D gives the route's history. Without the P1 meter check (\cref{sec:m-design}),
drift in the meter cannot be told from drift in the board.

\paragraph{Relation to MLPerf Tiny} These measurements do not follow MLPerf Tiny~\cite{mlperftiny},
which meters the core on its own channel over at least \SI{10}{\second} and reports the median of
five runs. Here the whole board is metered at its \SI{3.3}{\volt} rail, one inference per GPIO
window, so the figures compare across the rows and arms of this study and only loosely with
published MLPerf Tiny results.

\subsection{Verification against the reference kernels}\label{sec:m-verify}
Every INT8 accuracy was scored on the host and every hardware figure comes from a board, so I used
no hardware figure until the deployment behind it had been checked. The reference is TensorFlow
Lite's reference kernels run on the host with the same model file and input bytes, not the default
path, which does not reproduce them (\cref{sec:m-quant}). On the ESP32-S3 the six INT8 outputs had
to equal the reference element by element, with no tolerance, on 20 held-out photographs stratified
by class at row~3 and on the measurement image at rows 1, 2 and~4. On the RP2040 the test was exact
equality with the ESP32-S3's outputs for the same model and image.

A match says little where outputs sit on an INT8 limit ($-128$ or $127$), so each result carries its
count of saturated outputs, and two checks look inside the network. One reads the final dense layer
before softmax, through a probe that wraps that layer's kernel registration so no vendored source is
edited. The other compares a checksum of each of the 52 convolution and depthwise outputs at row~3
with the same layer run on the host. \Cref{sec:h-agree} gives the results and \cref{sec:h-tool} the
defects they turned up.

\paragraph{Addendum checks \rc{exploratory}} The addendum's checks (items C2, C4, C5, P2, P4 and
ESP4) appear beside the registered ones in \cref{sec:m-quant,sec:h-agree}.

\subsection{Use of AI tools in the research}\label{sec:m-ai}
From July 2026 I used Anthropic's Claude Sonnet 4.8 and~5 in the research. It helped me word the
analysis plan before it was registered. It also helped me debug the code and write sections of it,
including in the patch and tests for the ESP-NN kernels, the instrumentation that checks each board
against the reference kernels, and the host scorer on the reference kernels that re-scored every
INT8 model. I reviewed every change before using it, ran the code myself and checked its outputs,
and I take responsibility for it. No hardware figure was used until the deployment behind it matched
TensorFlow Lite's reference kernels exactly (\cref{sec:m-verify,sec:h-agree}), and all 430
registered INT8 models were rebuilt from their retained weights and re-scored, each reproducing its
recorded count (\cref{sec:m-quant}, items C2 and~C5). Both checks compare outputs with an independent
reference, so an error in the code would have shown up as a mismatch. The use of AI tools in preparing
the manuscript is declared separately, before the references.

\section{Registered results}\label{sec:results}
\begin{figure}[htbp]
\centering
\includegraphics[width=0.6\linewidth]{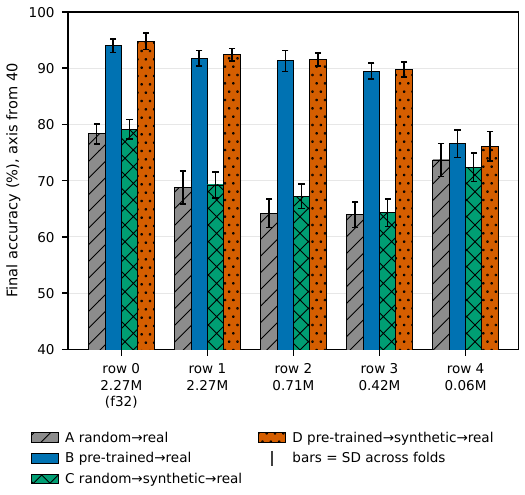}
\caption{Final test accuracy of the four arms at each row of the model ladder, at full
real data. Row~0 is shown in float32, the precision of its registered criterion; rows 1 to~4 are INT8.
Arms B and D at rows 3 and~4 are three-seed means; every other bar is seed~1. Bars show the
standard deviation across the ten folds, and the vertical axis starts at 40\%. At row~4, B and
D are B$'$ and D$'$. Rows 0 and~1 have the same parameter count. Descriptive: the registered
B-against-D tests are in \cref{tab:conf}.}
\label{fig:arms}
\end{figure}

For scale, \cref{fig:arms} shows all four arms at every row at full real data. Arms B and D start
from pre-trained weights and sit far above arms A and C at rows 0 to~3, and the gap between B and D,
which is all the registered tests concern, is too small to see at this scale. At row~4, where B$'$
and D$'$ start from Imagenette, the distance to A and C is much smaller. The figure is descriptive.

\subsection{Confirmatory tests}\label{sec:r-conf}
\Cref{tab:conf} lists the four registered tests, each paired over the ten folds and read as set out
in \cref{sec:m-stats}. Accuracies quoted beside a difference are means over the ten folds. The
addendum came back to all four, and its results are in \cref{sec:m-quant,sec:explore}; the values
here stay of record.

\paragraph{Anchor} On row~0, trained on 25\% of the photographs, arm~D finished below arm~B in
float32 by \SI{-1.068}{\pp} (SD \SI{0.843}{\pp}; 95\% CI \num{-1.945} to \num{-0.191}; $t_9 =
-2.756$, $p = 0.022$), 90.92\% against 91.99\%. Nine folds were negative and fold~7 tied exactly,
leaving none above zero, and the Wilcoxon test agrees ($W = 0$, $n = 9$, $p = 0.004$). With the
upper bound below $+1$ point, the anchor fails. The spread across folds was smaller than the pilot's
\SI{1.14}{\pp}, so the fail does not come from a wide interval. On this measurement the synthetic
stage left the largest model slightly worse off than ImageNet initialisation alone, which is
negative transfer relative to an already pre-trained baseline. The anchor rests on a single draw of
the synthetic stage, because one checkpoint serves all ten folds (\cref{sec:m-stats}); further draws
are in \cref{sec:explore}. Under the registration, a failed anchor means the effect is absent even
at this scale when real data is limited~\cite{yadav2026osf}. The ten fold values are in
Supplementary Table~S1.

\paragraph{Primary} At row~3 in INT8, with three-seed cell means, arm~D averaged 89.76\% and arm~B
89.45\%. The difference was \SI{+0.312}{\pp} (SD \SI{1.116}{\pp}; 95\% CI \num{-0.849} to
\num{+1.472}; $t_9 = 0.607$, $p = 0.559$, Holm-adjusted $p = 0.559$), with seven folds of ten above
zero; the Wilcoxon test gives $W = 18.5$ ($n = 10$, $p = 0.391$). No difference was detectable. The
upper bound, about four images on a 300-image fold, sits below the \SI{+1.6}{\pp} planning value, so
the registered bound holds: if the synthetic stage helps this model in INT8 at all, it helps by less
than the working-range gain recomputed from Bird et al. In float32 the same models gave
\SI{+0.044}{\pp} (\num{-0.292} to \num{+0.381}; five folds of ten). That is context, not a
registered test, but it means there was no float32 gain at row~3 for quantisation to remove.

\paragraph{Secondary} At row~4 both arms start from Imagenette weights, so the contrast isolates the
synthetic stage on that weaker start and cannot separate model size from starting point
(\cref{sec:m-arms}). In INT8 arm~D$'$ averaged 76.08\% and arm~B$'$ 76.53\%. The difference was
\SI{-0.445}{\pp} (SD \SI{0.348}{\pp}; 95\% CI \num{-0.807} to \num{-0.084}; $t_9 = -2.786$, $p =
0.021$), with one fold of ten above zero. As the smaller $p$-value in the Holm family it is doubled,
to 0.042, and the test rejects. The Wilcoxon test gives $W = 1$ ($n = 9$, one fold tied), $p =
0.008$. The difference is about 1.3 images per fold and has the same sign in float32
(\SI{-0.334}{\pp}; \num{-0.643} to \num{-0.025}), so quantisation did not produce it. Seed by seed
it is less firm: no single-seed interval excludes zero, and two of the three seeds carry the mean
(\cref{sec:r-seed}). I read it as a small, statistically detectable negative difference, which is
negative transfer relative to an already pre-trained baseline and nothing more.

The calibration bridge measures how much weaker the Imagenette start is. At row~3 it cost arm~B
\SI{17.09}{\pp} in INT8 (95\% CI \num{15.42} to \num{18.77}; all ten folds above zero), 72.36\%
against 89.45\%, and \SI{19.71}{\pp} in float32 (per fold in Supplementary Table~S1). Going from
row~3 to row~4 therefore changes the model and the starting point together. The registration's
reading rules had no branch for a significant negative result at row~4, so I add no interpretation
here; \cref{sec:r-seed} sets the result against run-to-run variance.

\paragraph{Difference-in-differences} At row~3 the difference-in-differences (\cref{eq:did}) was \SI{-0.267}{\pp}
(SD \SI{1.238}{\pp}; 95\% CI \num{-1.554} to \num{+1.020}; $t_9 = -0.469$, $p = 0.650$), with three
folds of ten above zero ($W = 18$, $n = 9$, $p = 0.633$). It is its own family and is not
Holm-adjusted. On average INT8 conversion cost arm~B \SI{2.35}{\pp} and arm~D \SI{2.08}{\pp}, and
the interval says it cost arm~D at most \SI{1.02}{\pp} more than arm~B, and arm~B at most
\SI{1.55}{\pp} more than arm~D. Filled in from the upper bound, the registered sentence would read
``quantisation removes at most 1.02 points of the transfer benefit''~\cite{yadav2026osf}. With no
float32 benefit at row~3, I read the interval only as a bound on how far INT8 moved the B-against-D
difference and do not interpret its sign. The row-4 difference-in-differences is descriptive and
appears in \cref{sec:explore}.

\begin{table}[htbp]
\centering
\caption{Registered confirmatory results. Differences are D$-$B (row~4:
D$'-$B$'$) in percentage points.}
\label{tab:conf}
\footnotesize
\begin{tabular}{@{}l l r c r c l@{}}
\toprule
Test & Contrast & Mean & 95\% CI & $p$ & \makecell{Folds\\$>0$} & Verdict \\
\midrule
Anchor & Row~0, float32, 25\% data & \num{-1.068} & \num{-1.945} to \num{-0.191} &
  0.022 & 0/10 & Fail \\
Primary & Row~3, INT8 & \num{+0.312} & \num{-0.849} to \num{+1.472} &
  0.559\textsuperscript{a} & 7/10 & \makecell[l]{No difference;\\excludes $+1.6$} \\
Secondary & Row~4, INT8 & \num{-0.445} & \num{-0.807} to \num{-0.084} &
  0.021\textsuperscript{a} & 1/10 & Negative \\
\makecell[l]{Difference-in-\\differences} & Row~3, float32 $-$ INT8 & \num{-0.267} &
  \num{-1.554} to \num{+1.020} & 0.650 & 3/10 & No difference \\
\bottomrule
\end{tabular}
\par\smallskip\raggedright\footnotesize\noindent\textit{Notes:} $p$-values are from
the corrected resampled $t$-test (\cref{eq:nb}; 9 degrees of freedom, two-sided) and are unadjusted.
\textsuperscript{a}Holm across the primary and secondary only: 0.559 and 0.042. The
anchor has one seed per cell; the other three tests use three-seed cell means. The
difference-in-differences is its own family. Exact Wilcoxon $p$-values, in table order:
0.004, 0.391, 0.008, 0.633. The anchor's registered criterion is float32; its INT8
values are measured and exploratory (\cref{sec:explore}).
\end{table}

\subsection{Registered trend regression}\label{sec:r-trend}
The registered optional regression asks whether the gap between B and D depends on model size
(\cref{sec:m-stats}). In float32, over rows 0 to~4, the mean slope was \SI{-0.055}{\pp} per decade
of parameters (SD \num{0.810}; 95\% CI \num{-0.896} to \num{+0.787}; $t_9 = -0.147$, $p = 0.886$),
with five of the ten fold slopes above zero. In INT8, over rows 1 to~4, it was \SI{+0.413}{\pp} per
decade (SD \num{1.111}; \num{-0.742} to \num{+1.567}; $t_9 = 0.809$, $p = 0.440$), with seven of ten
above zero. Neither interval excludes zero: in both precisions there is no detectable dependence on
model size. Taken one row at a time the seed-1 differences agree: none of the nine per-row intervals
excludes zero, and their means run from \SI{-0.633}{\pp} to \SI{+0.801}{\pp}. Supplementary Table~S2
lists them and Supplementary Figure~S1 plots them against parameter count.

The null does not show that the line is flat. The ladder spans only 1.57 decades of parameters, and
over that span the float32 interval still allows the difference to change by about $-1.4$ to
$+1.2$~\si{\pp} between the smallest model and the largest, and the INT8 interval by about $-1.2$ to
$+2.5$~\si{\pp}. Because every row enters at seed~1, the row-3 and row-4 inputs are not the
confirmatory values: \SI{-0.135}{\pp} against \SI{+0.312}{\pp} at row~3 and \SI{-0.067}{\pp} against
\SI{-0.445}{\pp} at row~4 in INT8. The addendum's three-replicate refit is in \cref{sec:explore}.

\subsection{Run-to-run spread and the paired difference}\label{sec:r-seed}
At rows 3 and~4, arms B and D have three runs per fold, and how far those runs scatter is this
pipeline's run-to-run variance. \Cref{tab:spread} gives it pooled over folds and arms by root mean
square (\cref{sec:m-stats}): in INT8, \SI{1.21}{\pp} at row~3 and \SI{1.36}{\pp} at row~4, about
four images on a 300-image fold. The confirmatory tests compare three-seed cell means paired by
fold, though, and the spread that matters for them is $\sigma_d$, \SI{1.116}{\pp} at row~3 and
\SI{0.348}{\pp} at row~4. If the arms were independent, the registered relation $\sigma_d \approx
\sqrt{\tfrac{2}{3}\sigma^2_{\mathrm{seed}} + \sigma^2_{\mathrm{int}}}$~\cite{yadav2026osf} with no
fold-by-arm term would predict \SI{0.985}{\pp} and \SI{1.111}{\pp}. The two rows miss in opposite
directions. At row~3 the measured value is a little higher, because the arms' INT8 cell means are
only moderately correlated across folds ($r = 0.675$, against 0.954 in float32). At row~4 it is
under a third of the prediction, because B$'$ and D$'$ move almost together ($r = 0.993$ in INT8): a
fold that is hard for one arm is hard for the other, and pairing cancels it.

That is why the row-4 difference can be small and still be resolved. At \SI{-0.445}{\pp} it is a
third of the pooled run-to-run standard deviation. Of the 60 pairs of runs that share an arm and a
fold at row~4, 50 differ by more than that, and the median gap is \SI{1.67}{\pp}, five images. By
that yardstick the effect sits well inside run-to-run variance, yet its interval excludes zero
because the test is made against the paired spread. ``Inside run-to-run variance'' and ``no effect''
are separate claims, and at row~4 only the first holds. The row-3 primary, \SI{+0.312}{\pp}, is 0.26
of the pooled standard deviation and its interval includes zero. Muralidharan et al.\ put no number
on the run-to-run variation they compare against~\cite{muralidharan2026}, and their contrast is
arm~B against arm~A (\cref{sec:rel-scale}), so the row-4 result neither confirms nor contradicts
theirs.

Averaging over seeds matters as much as pairing. One seed at a time, the INT8 difference at row~3 is
\num{-0.135}, \num{+0.434} and \SI{+0.635}{\pp} (paired standard deviations 1.85, 1.27 and
\SI{1.73}{\pp}), and at row~4 \num{-0.067}, \num{-0.836} and \SI{-0.433}{\pp} (0.77, 1.09 and
\SI{0.67}{\pp}). Every single seed is noisier than the three-seed mean, and at both rows the three
values span \SI{0.77}{\pp}. These per-seed figures are descriptive and are in Supplementary
Table~S2. The trend regression and the data-fraction sweep use seed~1 only, so their per-row values
carry spread of this size. The addendum's measure of how much of it comes from GPU non-determinism
is in \cref{sec:explore}.

\begin{table}[htbp]
\centering
\caption{Run-to-run spread and paired spread at rows 3 and~4, full real data.
Differences are D$-$B (row~4: D$'-$B$'$).}
\label{tab:spread}
\footnotesize
\begin{tabular}{@{}c l r r r r r r r r@{}}
\toprule
 & & \multicolumn{3}{c}{Run-to-run SD (pp)} & & & & & \\
\cmidrule(lr){3-5}
Row & Precision & B & D & \makecell[r]{Pooled\\(RMS)} & \makecell[r]{Predicted\\$\sigma_d$} &
  \makecell[r]{Measured\\$\sigma_d$} & $r$ & \makecell[r]{Mean\\$\Delta$} &
  \makecell[r]{$|\Delta|$ /\\pooled} \\
\midrule
3 & Float32 & 1.109 & 0.825 & 0.977 & 0.798 & 0.324 & 0.954 & \num{+0.044} & 0.05 \\
3 & INT8 & 1.330 & 1.069 & 1.206 & 0.985 & 1.116 & 0.675 & \num{+0.312} & 0.26 \\
4 & Float32 & 1.234 & 1.207 & 1.221 & 0.997 & 0.297 & 0.994 & \num{-0.334} & 0.27 \\
4 & INT8 & 1.282 & 1.436 & 1.361 & 1.111 & 0.348 & 0.993 & \num{-0.445} & 0.33 \\
\bottomrule
\end{tabular}
\par\smallskip\raggedright\footnotesize\noindent\textit{Notes:} Run-to-run SD is the root
mean square, over the ten folds, of the standard deviation across seeds 1 to~3; the
pooled value combines the two arms by root mean square. Predicted $\sigma_d$ is
$\sqrt{2/3}\times$ pooled, the registered relation with no fold-by-arm term. Measured
$\sigma_d$ is the standard deviation of the ten paired differences of three-seed cell
means, and $r$ is the correlation of the two arms' cell means across folds. All values were
recomputed from the image counts in the results file.
\end{table}

\section{Data-fraction sweep}\label{sec:fraction}
\begin{figure}[htbp]
\centering
\includegraphics[width=\linewidth]{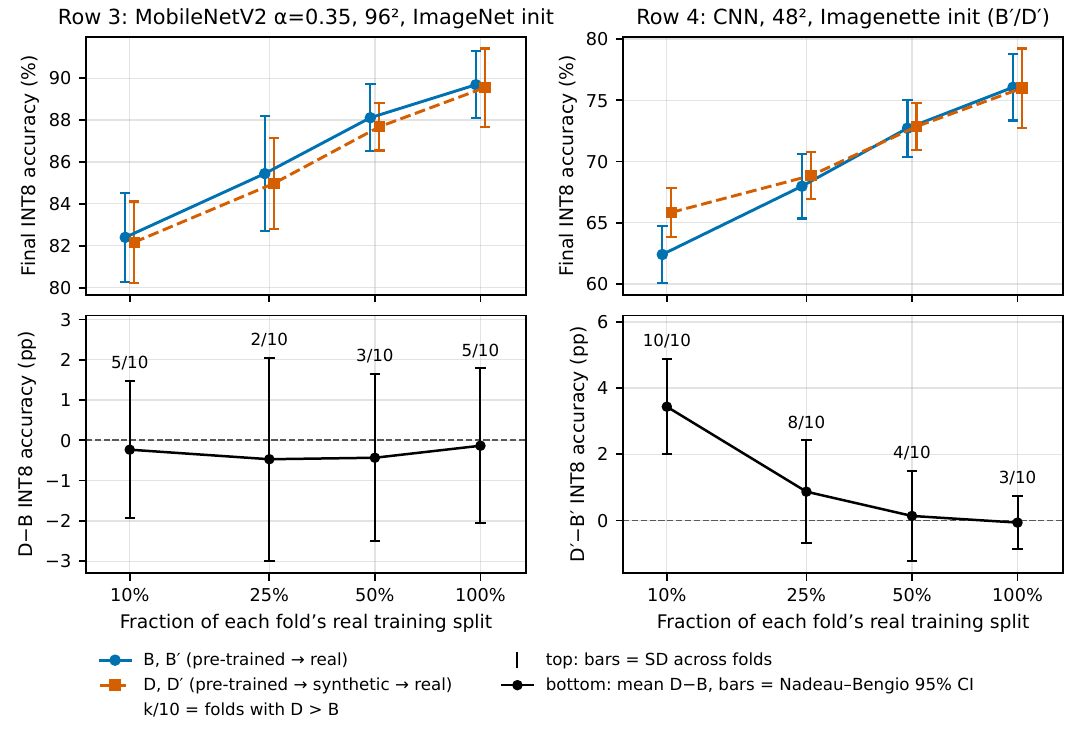}
\caption{INT8 accuracy against the share of real training data, at row~3 (left;
MobileNetV2, width 0.35, ImageNet start) and row~4 (right; CNN, Imagenette start, arms
B$'$ and D$'$). Top: mean accuracy of each arm over the ten folds, with bars showing
the standard deviation across folds. Bottom: mean of the ten paired differences, D$-$B
(row~4: D$'-$B$'$), with bars showing the corrected 95\% interval; $k$/10 is the number
of folds with D above B. The horizontal axis is logarithmic. Seed~1 at every fraction.
The representative set for INT8 conversion is drawn from each fold's full training
split and held fixed across fractions, so only the training data varies. Descriptive:
no test was registered on this sweep.}
\label{fig:fraction}
\end{figure}

I repeated arms B and D at rows 3 and~4 with 10, 25 and 50\% of each fold's real training
photographs, over the same ten folds with seed~1. The full-data point is the seed-1 run from the
main grid, not the three-seed confirmatory value (\cref{sec:r-trend}). B and D at a given fold and
fraction train on the same photographs, and the representative set is held fixed across fractions
(\cref{sec:m-quant}). No test was registered on these runs, so I give no $p$-values for them.
\Cref{tab:fraction} lists all sixteen cells, \cref{fig:fraction} plots them, and Supplementary
Table~S1 has the per-fold accuracies.

At row~4 with 10\% of the photographs, about 270 per fold, arm~D$'$ reached 65.84\% in INT8 and
arm~B$'$ 62.40\%. The difference was \SI{+3.438}{\pp} (SD \SI{1.388}{\pp}; 95\% CI \num{+1.996} to
\num{+4.881}), and all ten folds were above zero, by 6 to 19 images. Float32 gives almost the same,
\SI{+3.338}{\pp} (\num{+1.697} to \num{+4.979}; ten of ten). Both arms at this row start from
Imagenette weights, though, and the registered design cannot tell a scarce-data effect from a
weak-start effect, since the registered bridge ran at full data only. The gain is gone by 25\%:
\SI{+0.868}{\pp} (\num{-0.684} to \num{+2.420}; eight folds of ten), then \SI{+0.134}{\pp} at 50\%
and \SI{-0.067}{\pp} at full data, none of them excluding zero. Cutting from full data to 10\% cost
B$'$ \SI{13.66}{\pp} and D$'$ \SI{10.15}{\pp} in INT8.

Row~3 is flat. No interval at any fraction excludes zero in either precision, and the INT8 means run
from \SI{-0.470}{\pp} to \SI{-0.135}{\pp}, with two to five folds of ten above zero. Going from full
data to 10\% cost the ImageNet-initialised arms almost the same, 7.28 points for B and 7.38 for D.
The intervals reach 1.3 to 2.5 points either side of their means, so a small effect at row~3 is not
ruled out in either direction, but nothing like the row-4 pattern appears.

\paragraph{Addendum: replicates \rc{exploratory}} The addendum ran seeds 925 and~926 for every cell
below full data and averaged them with the registered seed-1 run (item C8). At row~4 with 10\% of
the photographs, D$'-$B$'$ was \SI{+2.715}{\pp} (95\% CI \num{+1.604} to \num{+3.826}; ten folds of
ten), against \SI{+3.438}{\pp} on seed~1. The two new seeds alone gave \SI{+2.354}{\pp}
(\num{+1.149} to \num{+3.558}; ten of ten), and float32 \SI{+2.625}{\pp} (\num{+1.348} to
\num{+3.903}). At 25 and 50\% the row-4 intervals include zero, and at row~3 no interval at any
fraction excludes it; from one seed to another the row-3 differences swing by up to \SI{1.5}{\pp}.
Supplementary Table~S2 sets every three-replicate cell beside its registered seed-1 value.

\begin{figure}[htbp]
\centering
\includegraphics[width=\linewidth]{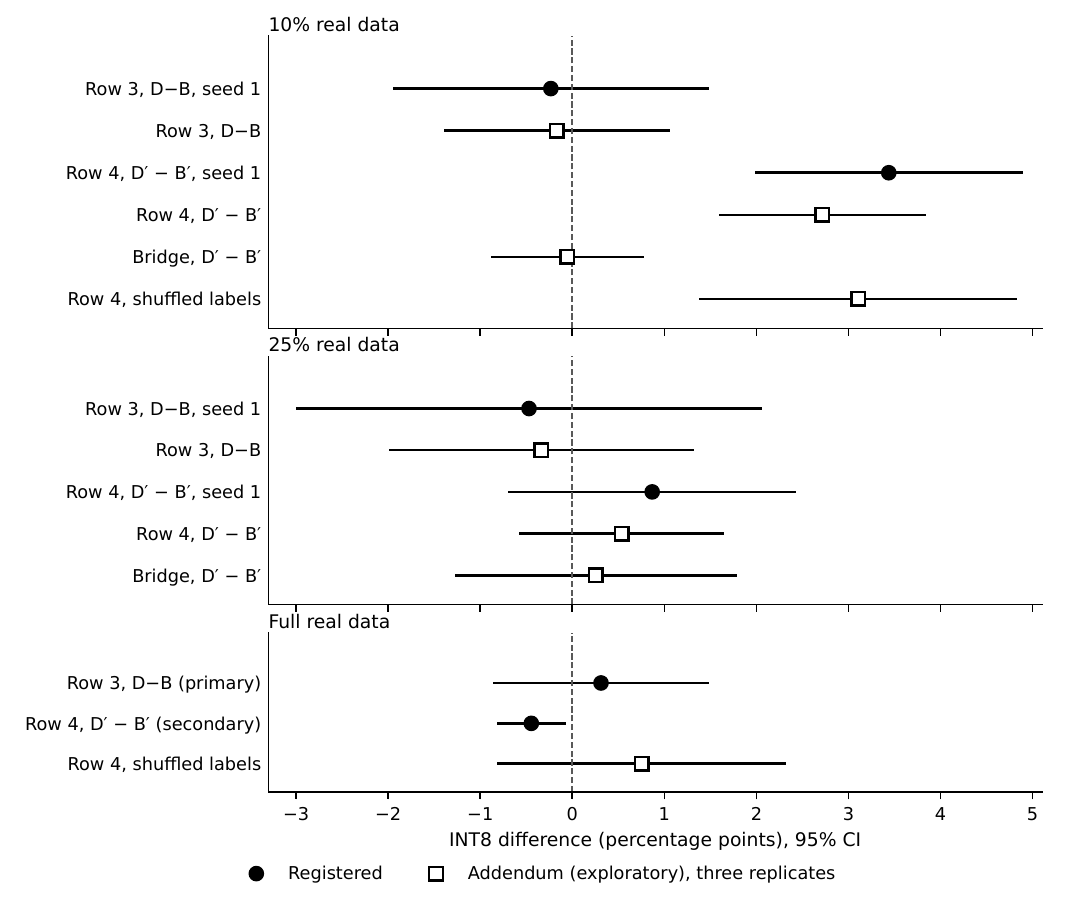}
\caption{INT8 differences with reduced and full real data, with the addendum's controls beside
the registered values. Filled circles are registered: the seed-1 sweep values, which are
descriptive registered measurements, and the primary and secondary tests at full data. Open
squares are exploratory addendum results over three replicates. Row~3 is MobileNetV2 at width
0.35 from ImageNet weights; row~4 is the CNN, arms B$'$ and D$'$, from Imagenette weights; the
bridge is the row-3 MobileNetV2 from Imagenette weights; ``shuffled labels'' is row~4 with the
labels of the rendered images shuffled. Bars are the corrected 95\% intervals over the ten
folds. Values are in \cref{tab:fraction,tab:conf} and Supplementary Table~S2.}
\label{fig:controls}
\end{figure}

\paragraph{Addendum: two controls \rc{exploratory}} Two items asked where the row-4 gain comes from
(\cref{fig:controls}). The first gave the calibration bridge a synthetic stage (item C9): the row-3
MobileNetV2 from its Imagenette weights (\texttt{init\_source='imagenette'}) ran as B$'$ and D$'$ at
10 and 25\% with three seeds. If the weak start were what lets the stage pay off, this model should
behave like row~4. It behaved like row~3: D$'-$B$'$ was \SI{-0.056}{\pp} at 10\% (\num{-0.872} to
\num{+0.760}; four folds of ten) and \SI{+0.256}{\pp} at 25\% (\num{-1.263} to \num{+1.775}; five of
ten). At 10\% the bridge sat \SI{2.771}{\pp} below row~4 (\num{-4.194} to \num{-1.347}; no fold
above zero) and \SI{0.111}{\pp} above row~3 (\num{-0.916} to \num{+1.138}). The Imagenette start
cost arm~B about \SI{20}{\pp} at both fractions, so the start was weak here as well.

The second control kept row~4's synthetic stage but shuffled the labels across the rendered images,
keeping the class counts (item C10). With 10\% of the photographs the shuffled-label D$'$ still
finished \SI{+3.104}{\pp} above B$'$ (\num{+1.394} to \num{+4.813}; ten of ten), and true-label
minus shuffled-label D$'$ was \SI{-0.389}{\pp} (\num{-1.488} to \num{+0.710}; three of ten). At full
data the shuffled-label difference was \SI{+0.756}{\pp} (\num{-0.800} to \num{+2.312}; six of ten)
and true minus shuffled \SI{-1.201}{\pp} (\num{-2.842} to \num{+0.441}; three of ten). The sweep has
no full-data cell, so the true-label side there is the registered three-seed cell behind the
secondary test, a stand-in the addendum did not name. The row-0 arm of this control was not run
(\cref{sec:m-design}).

\paragraph{Reading the sweep with the addendum} The registered seed-1 values stay the measurements
of record. The row-4 gain at 10\% replicates: it is smaller over three replicates than on seed~1,
every fold stays above zero, and it holds without the registered run. It is the only contrast to
survive the addendum's screen (raw $p = 0.0004$, adjusted $p = 0.006$; \cref{tab:bh}), which is a
screen and not a test. The controls narrow the cause: a MobileNetV2 with the same kind of weak start
gets nothing from the synthetic stage, so the weak start does not explain the gain on its own. The
gain also survives shuffled labels, which rules out label content. That leaves something in the
rendered images themselves or in the extra training, acting on row~4's model. The bridge sees
\SI{96}{px} inputs and row~4 \SI{48}{px}, and nothing here separates the model from its input
resolution; the mechanism stays open.

\begin{table}[htbp]
\centering
\caption{Data-fraction sweep at rows 3 and~4, seed~1. Accuracies are INT8 means over the
ten folds; differences are D$-$B (row~4: D$'-$B$'$) in percentage points.}
\label{tab:fraction}
\footnotesize
\begin{tabular}{@{}c r r r r c c r c c@{}}
\toprule
 & & \multicolumn{5}{c}{INT8} & \multicolumn{3}{c}{Float32} \\
\cmidrule(lr){3-7}\cmidrule(l){8-10}
Row & \makecell[r]{Real\\data} & B & D & Mean & 95\% CI & \makecell{Folds\\$>0$} &
  Mean & 95\% CI & \makecell{Folds\\$>0$} \\
\midrule
3 & 10\% & 82.402 & 82.169 & \num{-0.233} & \num{-1.933} to \num{+1.467} & 5/10 &
  \num{-0.270} & \num{-2.273} to \num{+1.733} & 4/10 \\
3 & 25\% & 85.445 & 84.975 & \num{-0.470} & \num{-2.989} to \num{+2.049} & 2/10 &
  \num{-0.703} & \num{-2.490} to \num{+1.085} & 3/10 \\
3 & 50\% & 88.112 & 87.679 & \num{-0.433} & \num{-2.503} to \num{+1.637} & 3/10 &
  \num{-0.135} & \num{-1.773} to \num{+1.503} & 5/10 \\
3 & 100\% & 89.685 & 89.550 & \num{-0.135} & \num{-2.053} to \num{+1.784} & 5/10 &
  \num{-0.033} & \num{-1.324} to \num{+1.258} & 4/10 \\
\midrule
4 & 10\% & 62.404 & 65.842 & \num{+3.438} & \num{+1.996} to \num{+4.881} & 10/10 &
  \num{+3.338} & \num{+1.697} to \num{+4.979} & 10/10 \\
4 & 25\% & 67.980 & 68.848 & \num{+0.868} & \num{-0.684} to \num{+2.420} & 8/10 &
  \num{+0.534} & \num{-1.052} to \num{+2.120} & 8/10 \\
4 & 50\% & 72.721 & 72.855 & \num{+0.134} & \num{-1.226} to \num{+1.493} & 4/10 &
  \num{+0.367} & \num{-0.955} to \num{+1.689} & 5/10 \\
4 & 100\% & 76.062 & 75.995 & \num{-0.067} & \num{-0.864} to \num{+0.730} & 3/10 &
  \num{+0.200} & \num{-0.516} to \num{+0.916} & 5/10 \\
\bottomrule
\end{tabular}
\par\smallskip\raggedright\footnotesize\noindent\textit{Notes:} Real data is the share
of each fold's real training photographs; the test fold is always whole. Intervals are
the corrected 95\% intervals over the ten folds. One seed per cell: the 100\% rows
are the seed-1 values and differ from the three-seed confirmatory results
(\cref{tab:conf}). Descriptive throughout: no test was registered on these cells.
\end{table}

\section{Exploratory analyses}\label{sec:explore}
Nothing in this section was a registered test, and nothing here is Holm-adjusted. The results that
came before the addendum carry no $p$-values. Supplementary Table~S2 gives the values in this
section from before the addendum, apart from the three-fold screens and the resolution sweep
(Supplementary Note~E), and those of items C7 to C11, with their intervals, including all
60 pairwise contrasts between the four arms at every row in both precisions; Supplementary
Table~S6 gives the outcomes of items C6, C12 and C14. Two of the 60 are the primary and secondary
tests, and I looked at the other 58 without a test. In 44 of the 60 the interval excludes zero: 40
pit an arm that starts from pre-trained weights against one that starts from random weights, and
the other four are C$-$A at row~2 and D$'-$B$'$ at row~4, each in both precisions. The addendum's
declared accuracy contrasts carry raw $p$-values, and \cref{tab:bh} at the end of the section gives
each with its Benjamini--Hochberg adjustment.

\paragraph{Row~0 in INT8} INT8 was measured for every row-0 run and is exploratory. With 25\% of the
photographs the INT8 difference was \SI{-0.534}{\pp} (95\% CI \num{-1.659} to \num{+0.591}; two
folds of ten above zero), the same sign as the float32 anchor and about half its size. With all the
photographs it was \SI{+1.034}{\pp} (\num{-0.377} to \num{+2.445}; eight of ten), against
\SI{+0.801}{\pp} in float32. Conversion cost row~0 very little, \SI{0.40}{\pp} for arm~B and
\SI{0.17}{\pp} for arm~D on one seed, against 2.35 and \SI{2.08}{\pp} for the three-seed cells at
row~3.

\paragraph{Row~0 across the two data fractions} From 25\% to full data D$-$B moved by
\SI{+1.869}{\pp} in float32 (\num{-0.122} to \num{+3.860}) and \SI{+1.568}{\pp} in INT8, neither
interval excluding zero. At 25\% an epoch is 22 batches against 85, so data and optimiser steps fall
together. Arm~D trailed arm~B at 40 of the 41 epochs after the first, and the anchor's deficit on
that draw was not an accident of the final epoch (Supplementary Note~E).

\paragraph{Addendum: four draws of the anchor \rc{exploratory}} After the anchor had been read I
trained a second synthetic checkpoint (seed 911); against the registered arm~B it gave
\SI{-0.201}{\pp} in float32 (\num{-1.878} to \num{+1.475}), neither repeating the fail nor
contradicting it (Supplementary Table~S2). The addendum trained two more checkpoints (seeds 921
and~922) and two more arm-B runs (seeds 923 and~924), so each of four arm-D draws could be set
against a three-replicate arm-B mean at the same fold (item C7). The four draws gave \num{-0.945},
\num{-0.078}, \num{-0.880} and \SI{-0.513}{\pp}, the second being seed 911. The registered draw
comes out at \SI{-0.945}{\pp} here, not \SI{-1.068}{\pp}, because the seed-1 arm-B run was the
highest of the three. Pooled with fold and draw as crossed random effects, the float32 difference
was \SI{-0.604}{\pp} (95\% CI \num{-1.239} to \num{+0.031}), and the draws spread with a standard
deviation of 0.399, about what fold noise alone gives (0.396). Averaged over draws within each fold,
as declared for the screen, it was \SI{-0.604}{\pp} (\num{-1.233} to \num{+0.025}; raw $p = 0.058$,
adjusted $p = 0.232$). In INT8 the pooled difference was \SI{-0.212}{\pp} (\num{-1.082} to
\num{+0.658}). A new draw also reshuffles the order of the real-data stage, because both follow the
run seed, and nothing here separates the two.

\paragraph{Reading the anchor with the addendum} The registered fail is unchanged, and every draw
came out negative. Against more draws and more arm-B runs, though, the deficit is about
\SI{0.6}{\pp} and its interval reaches just above zero, so part of the registered estimate belongs
to the particular checkpoint and arm-B run it used.

\paragraph{Other follow-ups} Three-fold screens of other synthetic-stage settings at rows 0 and~1,
and a sweep of input resolution at row~3, alter no registered result; Supplementary Note~E gives
both.

\paragraph{Synthetic images without pre-training (C$-$A)} Arms A and~C both start from random
weights, so C$-$A shows what the rendered images do for an untrained model. At full data and seed~1
only row~2's interval excludes zero (float32 \SI{+3.742}{\pp}, \num{+1.453} to \num{+6.030}; INT8
\SI{+3.039}{\pp}), against 15.7 to \SI{28.1}{\pp} for an ImageNet start (B$-$A) at rows 0 to~3. One
row in five is not a pattern, and the comparison partly measures speed of convergence~\cite{he2019};
arm~C also gets two extra epochs and settings chosen on arm~D. Supplementary Table~S2 gives every
row.

\paragraph{Difference-in-differences at row~4} Registered as descriptive, it came to
\SI{+0.111}{\pp} (\num{-0.414} to \num{+0.637}), with four folds of ten above zero and one at zero;
INT8 cost arm~B$'$ \SI{0.22}{\pp} and arm~D$'$ \SI{0.33}{\pp}. The registration makes it
interpretable only if row~4's float32 benefit is non-zero~\cite{yadav2026osf}, and that difference
was a deficit (\SI{-0.334}{\pp}; \cref{sec:r-conf}), so I leave the number uninterpreted. The seed-1
differences-in-differences at rows 0 to~3 all have intervals that include zero.

\paragraph{Addendum: the trend with three replicates \rc{exploratory}} The addendum added seeds 936
and~937 at rows 0 to~2, so that every row has three replicates in arms B and~D at full data (item
C11). The fit declared for the screen, INT8 over rows 0 to~4, gave \SI{+0.513}{\pp} per decade of
parameters (95\% CI \num{+0.088} to \num{+0.937}; raw $p = 0.023$, adjusted $p = 0.124$). The same
fit on seed~1 alone gives \SI{+0.608}{\pp} (\num{-0.098} to \num{+1.314}); the estimate barely moved
and the interval narrowed as rows 0 to~2 became less noisy. The slope rests mostly on row~4: over
rows 0 to~3 it is \SI{+0.262}{\pp} (\num{-1.204} to \num{+1.727}). The two fits that match the
registered ones, INT8 over rows 1 to~4 (\SI{+0.438}{\pp}; \num{-0.066} to \num{+0.942}) and float32
over rows 0 to~4 (\SI{+0.250}{\pp}; \num{-0.085} to \num{+0.584}), both include zero, and no per-row
interval at rows 0 to~2 excludes it. Supplementary Table~S2 gives the per-row values and all five
fits, each beside its seed-1 fit where there is one, and Supplementary Figure~S3 plots the
per-row values with the fit over rows 0 to~4.

\paragraph{Reading the trend with the addendum} The registered result, no detectable dependence on
model size, still holds. The exploratory refit has a positive slope that excludes zero before
adjustment and not after, carried mostly by the small row-4 deficit, and I leave the question open.

\paragraph{Addendum: a second training recipe \rc{exploratory}} The addendum re-ran rows 3 and~4 at
full data, arms B and~D (B$'$ and D$'$), with seeds 938 to 940, under a recipe that keeps the flip
and adds a random crop after resizing to 115\%, rotation of up to $\pm 10$\textdegree{} and changes
of up to $\pm 20$\% in brightness and contrast, in both stages (item C12). It cost accuracy in every
arm, by 3.6 and \SI{5.9}{\pp} for B and~D at row~3 and by 22.9 and 22.0 points for B$'$ and D$'$ at row~4. Here it is a
worse recipe, not a better-trained check. At row~3 the INT8 difference turned negative,
\SI{-1.915}{\pp} (\num{-3.470} to \num{-0.359}; one fold of ten above zero; raw $p = 0.021$,
adjusted $p = 0.124$), against the registered \SI{+0.312}{\pp}, and float32 agrees
(\SI{-1.480}{\pp}; \num{-2.055} to \num{-0.904}). At row~4 it was \SI{+0.468}{\pp} (\num{-0.560} to
\num{+1.496}; six of ten), against the registered \SI{-0.445}{\pp}. Between the two recipes the
contrast moved by \SI{-2.226}{\pp} at row~3 (\num{-4.208} to \num{-0.244}) and by \SI{+0.913}{\pp}
at row~4 (\num{-0.190} to \num{+2.017}).

\paragraph{Reading the recipe result} The registered primary and secondary hold for the recipe they
were run under. With heavier augmentation the row-3 null becomes a deficit and the row-4 deficit
disappears, a dependence on the recipe and not a correction to either registered result.

\paragraph{Addendum: row~3 without the tuning folds \rc{exploratory}} The two decision experiments
ran on folds 0 to~2, which are also test folds for the primary (\cref{sec:m-train}). Over folds 3
to~9, with the correction for seven folds, D$-$B at row~3 was \SI{0.000}{\pp} in float32
(\num{-0.448} to \num{+0.448}) and \SI{+0.128}{\pp} in INT8 (\num{-1.455} to \num{+1.710}; item
C14). Both are null, but with seven folds the INT8 upper bound no longer excludes the \SI{+1.6}{\pp}
planning value, so the registered ten-fold bound is a narrow one. The minimum detectable effects say
the same. At 80\% power and a two-sided level of 0.05 they are 1.22 points for the anchor, 1.62 for
the primary, 0.50 for the secondary and 1.79 for the difference-in-differences, so the primary was
built to detect about the planning value and nothing smaller. The addendum named $\sigma_d$ and
$\sigma_{\mathrm{DiD}}$ as the spreads. For the anchor and the secondary I used their own observed
spreads (0.843 and \SI{0.348}{\pp}), a departure from that wording; at $\sigma_d$ both would be
1.62.

\paragraph{Addendum: same-seed spread \rc{exploratory}} One cell, row~3, arm~D, fold~0, seed~1, was
run five times with TensorFlow's deterministic ops off and five times with them on (item C6). Off,
float32 accuracy ranged from 272 to 276 images of 300 and INT8 from 263 to 268; on, every run gave
273 and 269. So the spread at a fixed seed is GPU non-determinism, against a spread between seeds of
about 1.0 to \SI{1.2}{\pp} in the same cell, and deterministic ops made training about 2.3 times
slower.

\paragraph{Addendum: per-class results and cost \rc{exploratory}} The re-scoring of every registered
model (item C5; \cref{sec:m-quant}) saved per-image predictions, and item C14 used them for
per-class results. Only row~4 at 10\% shows a clear
macro-F1 gain for the synthetic stage, \SI{+3.77}{\pp} on seed~1, with recall up in all six classes;
at full data, on the three-seed cell means, row~4 is slightly negative (\SI{-0.41}{\pp}). The synthetic stage is cheap: the 14
checkpoints behind the registered runs took \SI{911.8}{\second} on a T4 in all, about
\SI{4.3}{\second} for each arm-C or arm-D run that shares them.

\paragraph{Benjamini--Hochberg screen} \Cref{tab:bh} lists the 16 accuracy contrasts the addendum
declared, with raw and adjusted $p$-values. Only row~4 at 10\% survives an adjusted level of 0.05.
The screen says which differences deserve a direct test; it does not test them.

\begin{table}[htbp]
\centering
\caption{The addendum's Benjamini--Hochberg screen: 16 exploratory accuracy contrasts, false
discovery rate 0.05, two-sided.}
\label{tab:bh}
\footnotesize
\begin{tabular}{@{}r l l r r@{}}
\toprule
Rank & Item & Contrast & Raw $p$ & Adjusted $p$ \\
\midrule
1 & C8 & Row~4, 10\%, INT8 D$'-$B$'$ & 0.0004 & 0.006 \\
2 & C12 & Row~3, second recipe, INT8 D$-$B & 0.021 & 0.124 \\
3 & C11 & Trend slope, INT8, rows 0 to~4 & 0.023 & 0.124 \\
4 & C7 & Row~0, four draws, float32 D$-$B & 0.058 & 0.232 \\
5 & C10 & Row~4, 100\%, true $-$ shuffled D$'$ & 0.132 & 0.423 \\
6 & C8 & Row~4, 25\%, INT8 D$'-$B$'$ & 0.301 & 0.692 \\
7 & C12 & Row~4, second recipe, INT8 D$'-$B$'$ & 0.330 & 0.692 \\
8 & C8 & Row~3, 50\%, INT8 D$-$B & 0.346 & 0.692 \\
9 & C10 & Row~4, 10\%, true $-$ shuffled D$'$ & 0.444 & 0.790 \\
10 & C8 & Row~3, 25\%, INT8 D$-$B & 0.656 & 0.940 \\
11 & C9 & Bridge, 25\%, INT8 D$'-$B$'$ & 0.712 & 0.940 \\
12 & C8 & Row~3, 10\%, INT8 D$-$B & 0.764 & 0.940 \\
13 & C14 & Row~3, folds 3 to~9, INT8 D$-$B & 0.850 & 0.940 \\
14 & C9 & Bridge, 10\%, INT8 D$'-$B$'$ & 0.880 & 0.940 \\
15 & C8 & Row~4, 50\%, INT8 D$'-$B$'$ & 0.881 & 0.940 \\
16 & C14 & Row~3, folds 3 to~9, float32 D$-$B & 1.000 & 1.000 \\
\bottomrule
\end{tabular}
\par\smallskip\raggedright\footnotesize\noindent\textit{Notes:} Exploratory. Raw $p$-values
come from the corrected paired $t$-test (\cref{sec:m-stats}); C14 uses the correction for seven
folds. Adjusted values are a screen for which differences deserve a direct test, not tests, and
none of these contrasts enters the registered Holm or difference-in-differences families. An
adjusted value is the smallest of $p_{(j)} \times 16/j$ over ranks $j \geq i$, capped at~1. The
row-16 values are 0.9996 before rounding.
\end{table}

\section{Hardware cost ladder}\label{sec:hardware}
\begin{figure}[htbp]
\centering
\includegraphics[width=\linewidth]{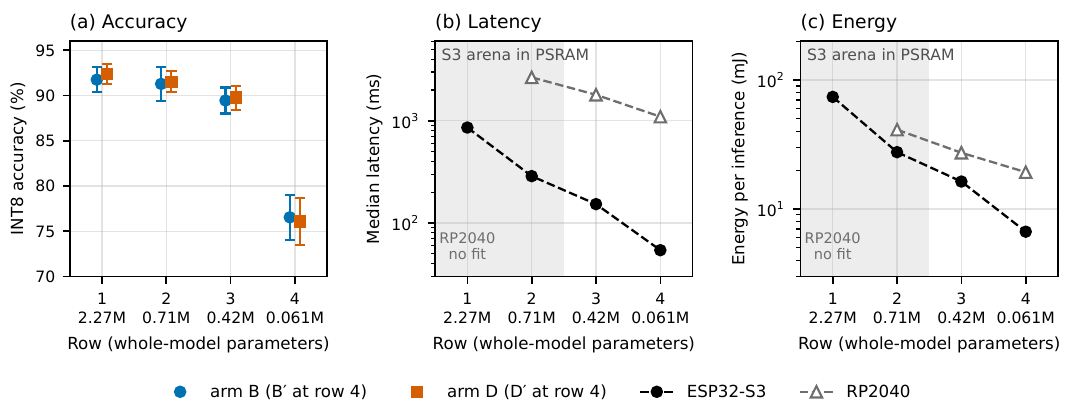}
\caption{Accuracy, latency and energy for the four deployed rows. (a)~INT8 test accuracy of
arms B and~D (B$'$ and D$'$ at row~4), mean and standard deviation over ten folds, on the host's
default kernel path; rows 3 and~4 are three-seed means. (b)~Median latency: ESP32-S3 at
\SI{240}{\mega\hertz} on the ladder build (\texttt{main\_T36.cc}), RP2040 at
\SI{133}{\mega\hertz} on its boundary-test build; row~1 does not fit in the RP2040's flash.
(c)~Idle-subtracted energy per inference (\cref{sec:h-energy}); RP2040 energy is a
post-registration addition. Shading marks the ESP32-S3 rows whose arena sits in PSRAM. Filled
markers are the ESP32-S3, open markers the RP2040. Row~4 on the ESP32-S3 includes ESP-NN 1.3.1's
debug heap check (\cref{sec:h-esp}). Row~0 is not shown.}
\label{fig:combined}
\end{figure}

\Cref{fig:combined} puts accuracy, latency and energy for the four deployed rows on one axis. Down
the ladder, arm~B's INT8 accuracy moves by about two points from row~1 to row~3 (91.8\% to 89.4\%)
and then falls by nearly 13 at row~4, while latency and energy fall with every step on both boards.
The rows are different models, so the figure does not vary one thing at a time.

\subsection{Device-versus-reference agreement}\label{sec:h-agree}
\Cref{tab:agree} lists every check of \cref{sec:m-verify}, with the number of outputs on an INT8
limit beside each result. On the ESP32-S3 every check was exact. On the RP2040 the outputs departed
from the reference.

\begin{table}[htbp]
\centering
\caption{Device-versus-reference agreement. Each device's INT8 outputs are compared
element by element with TensorFlow Lite's reference kernels run on the host, with no
tolerance.}
\label{tab:agree}
\small
\setlength{\tabcolsep}{4pt}
\begin{tabular}{@{}c l c c c c@{}}
\toprule
Row & Images & Tensor & \makecell{Values on an\\INT8 limit} &
\makecell{ESP32-S3\\exact} & \makecell{RP2040 exact\\(largest diff.)} \\
\midrule
3 & 20 & Output & 111/120 & 120/120 & \cref{tab:pico20} \\
3 & 20 & Pre-softmax\textsuperscript{a} & 0/120 & 120/120 & -- \\
\midrule
1 & 1 & Output & 6/6 & 6/6 & Does not fit \\
2 & 1 & Output & 2/6 & 6/6 & 4/6 (2) \\
3 & 1 & Output & 5/6 & 6/6 & 6/6 (0) \\
4 & 1 & Output & 0/6 & 6/6 & 0/6 (13) \\
\bottomrule
\end{tabular}
\par\smallskip\raggedright\footnotesize\noindent\textit{Notes:} Each row is the INT8
arm-B file (B$'$ at row~4) for fold~0, seed~1. The twenty row-3 images come from fold~0's
test split; the single image is the measurement image used for every hardware figure.
``Values on an INT8 limit'' counts reference outputs equal to $-128$ or $127$; a match
there shows little. Differences are in output steps. On the single image, both boards
predicted living room, the reference class, on every row where they ran. The ESP32-S3
results for rows 1, 2 and~4 are from the ladder lineage (\texttt{main\_T36.cc}); row~4
matched in both its SRAM and PSRAM builds, and row~3 in both ladder builds.
\textsuperscript{a}Output of the final dense layer, before softmax. Descriptive; the
criterion was set on the output tensor only.
\end{table}

\paragraph{ESP32-S3} At row~3 the check used 20 photographs from fold~0's test split, three per
class plus a fourth for forest and for living room, picked by a seeded rule; none is the measurement
image. Each was run 21 times, and all 120 output values matched the reference. That is weaker than
it sounds. Fifteen of the twenty reference vectors sit entirely on the limits, 111 of the 120
values, so only nine values on five images test anything finer than which class saturates. Before
softmax none of the 120 reference values sits on a limit; all twenty vectors differ, the values run
from $-123$ to 97, and the device matched all 120. On the measurement image, all 52 convolution and
depthwise checksums matched the reference in both the optimised and the plain-C ESP-NN builds, while
TensorFlow Lite's built-in kernels without the default delegate matched only 2 of the 52, so exact
agreement does tell kernel paths apart. The default host path gave identical vectors on 18 of the 20
images and differed by up to 15 steps on two unsaturated ones, with the same class throughout; the
board sits on the reference side of the split in \cref{sec:m-quant}.

Rows 1, 2 and~4 matched 6 of 6 on the measurement image, with the output constant over all 21
inferences, and row~4 gave bit-identical outputs in its SRAM and PSRAM builds. The rows carry
different weight. Row~4 has no output on a limit and is the strongest test, row~2 has two and row~1
all six. Row~2 was checked blind: I had not seen its device outputs when its reference values were
computed. Row~2 is also the only one of the three where the host paths disagree: the device matched 
the reference path, and neither the default path nor TensorFlow Lite's built-in kernels without the 
default delegate.

\paragraph{RP2040} Row~3 matched on all six outputs, row~2 on four (largest difference 2 steps) and
row~4 on none (largest difference 13). Counting by saturation changes the picture. All seven
reference outputs on a limit agreed, which shows nothing. Of the eleven that are not, three agreed,
all within two steps of the floor, and none of the eight further from a limit did. So the RP2040
build does not reproduce the reference arithmetic, and the rows that seem to agree do so because
their outputs are clamped. The departure is not an input fault (row~4's re-quantised input equals
the compiled-in header), not corruption (every output vector dequantises to probabilities that sum
to one) and not timing (the same outputs came back at 133 and \SI{200}{\mega\hertz}, with interrupts
on or masked).

\paragraph{Addendum: re-runs on the ESP32-S3 \rc{exploratory}} Repeated with logs kept (item ESP4),
row~3 again matched on all 20 photographs, before softmax and after. Rows 1 and~2 matched on one
image each in both tensors, and each added a check: row~1 now has a pre-softmax check, and row~2 was 
run on a fresh image whose reference values were hashed before the run. On that image the default
path and the built-in kernels without the default delegate are both far from the reference, and the
device matched the reference.

\paragraph{Addendum: where the RP2040 departs \rc{exploratory}} An RP2040 build that records a
checksum after every layer (item P2) was compared with per-layer reference checksums from the host
(item C4). At all three rows every layer matched up to the global average pool (MEAN), the first
operator to differ: operator~7 at row~4 and operator~62 at rows 3 and~2. The convolution, depthwise,
add and max-pooling layers before it are bit-identical, and the dense and softmax layers differ only
because their input already does. MEAN is TensorFlow Lite Micro's generic reduction kernel, not a
CMSIS-NN kernel, so the rebuild with reference kernels in place of CMSIS-NN (item P3) was not run.
What inside MEAN causes the departure was not tested and is open. Over the twenty row-3 photographs,
resized to each row's input (item P4; \cref{tab:pico20}), exact output vectors were 5, 18 and 17 of
20 at rows 4, 3 and~2, and the predicted class agreed on 18, 19 and 20. A second pass gave the same
outputs, every reference vector wholly on the limits matched, and agreement is weakest where the
outputs are not clamped.

\begin{table}[htbp]
\centering
\caption{RP2040 against the reference kernels on 20 photographs (addendum).}
\label{tab:pico20}
\footnotesize
\begin{tabular}{@{}c c c c@{}}
\toprule
Row & Exact output vectors & Same predicted class & Largest difference (steps) \\
\midrule
4 & 5/20 & 18/20 & 16 \\
3 & 18/20 & 19/20 & 15 \\
2 & 17/20 & 20/20 & 12 \\
\bottomrule
\end{tabular}
\par\smallskip\raggedright\footnotesize\noindent\textit{Notes:} Exploratory; item P4. INT8 arm-B
files
(B$'$ at row~4) for fold~0, seed~1, at \SI{133}{\mega\hertz}, on the twenty fold-0 photographs
used for the row-3 check in \cref{tab:agree}. A second pass gave identical outputs.
\end{table}

\paragraph{What the checks cover} The checks cover one deployed file per row, from fold~0 and
seed~1, on one board of each kind. The addendum's ESP32-S3 builds, row~0 and the plain-C builds
included, all matched the reference on the twenty photographs (\cref{sec:h-esp}). Nothing here says
the other cells of the grid would match.

\subsection{Defects and discrepancies in the toolchain}\label{sec:h-tool}
Making the boards pass the checks of \cref{sec:h-agree}, and the checks themselves, turned up
problems at several points between a trained model and a board (\cref{tab:toolchain}). Four are
defects in ESP-NN's optimised kernels for the ESP32-S3. One stopped row~3 running at all, and none
of the other three is reached by the deployed models. The table's fourth ESP-NN row is not a fault in the arithmetic: ESP-NN 1.3.1 walks
the heap three times per inference at row~4, a debug check that sits inside row~4's
latency and energy of record. Two more give no sign of trouble on their own,
the host's default path and the RP2040 build; only a comparison with the reference shows either.
None of them changes a registered result. Two runtime behaviours, input-buffer reuse and
non-reproducible conversion, are handled in \cref{sec:m-hw,sec:m-quant}.

\paragraph{The depthwise overrun} With ESP-NN 1.3.1's optimised kernels, row~3 crashed the board:
the tiled depthwise kernel wrote past the scratch buffer it had been given. I patched the kernel
locally, and with the patch the board matched the reference on all 52 per-layer checksums, in the
optimised build as in the plain-C one. Every ESP32-S3 figure of record uses the patched kernel.

\paragraph{Addendum: the overrun in a newer release \rc{exploratory}} The overrun does not recur in
the unpatched ESP-NN 1.4.1 release~\cite{espnn141} (item ESP2). The board did not crash, and row~3's
outputs and all 52 checksums matched the reference. A hardware watchpoint just past each of the 17
depthwise scratch buffers of the row-3 model never fired on 1.4.1 and fired on the first write past
the end on unpatched 1.3.1. With guard bytes after scratch buffers of exactly the declared size, all
68 depthwise layers of rows 0 to~3 ran clean on 1.4.1. Bisecting the upstream history put the fix in
commit \texttt{85961e4}, whose message does not mention the overrun.

\paragraph{Defects in the vendor kernels \rc{exploratory}} While checking the 1.4.1 release I also
tested ESP-NN's optimised ESP32-S3 kernels on shapes outside those my models use, and found three
further defects. With fixes for all three in place the board passed all 328 tests in my suite,
against 318 without them. I checked the files deployed at rows 1 to~4 against all three, and none of
the defects is reached by them, and no result in this paper depends on them. Row~0 has the same
operator types as row~1 at a larger input, but I did not check its file. One defect was in the
max-pool kernel, which ignored the lower bound of a fused activation when the channel count was 4
modulo 8. I reported it with a fix and regression tests, and the fix has been merged
upstream~\cite{espnnpr42}. I reported the other two privately to Espressif, which reproduced
both and has told me that fixes based on my patches will ship in ESP-NN 1.4.2. The same
random-shape tests on ESP-NN 1.3.1, the version of record, also found its softmax differing from ESP-NN's own
plain-C reference in 49 of 500 cases, and three depthwise cases, which I did not classify. The
deployed files are unaffected, since their outputs matched the reference kernels exactly
(\cref{sec:h-agree}).

\begin{table}[htbp]
\centering
\caption{Defects and discrepancies found in the toolchain between the trained model and the
boards.}
\label{tab:toolchain}
\footnotesize
\setlength{\tabcolsep}{3pt}
\begin{tabular}{@{}>{\raggedright\arraybackslash}p{3.3cm}
                  >{\raggedright\arraybackslash}p{2.6cm}
                  >{\raggedright\arraybackslash}p{3.7cm}
                  >{\raggedright\arraybackslash}p{3.0cm}
                  >{\raggedright\arraybackslash}p{2.3cm}@{}}
\toprule
Finding & How it showed & Effect on this paper & Status & Source \\
\midrule
\multicolumn{5}{@{}l}{\textit{ESP-NN optimised kernels, ESP32-S3}} \\
Out-of-bounds write in the tiled depthwise kernel, version 1.3.1 & Board crashed on row~3 &
  Patched locally; every ESP32-S3 figure of record uses the patch & Absent from 1.4.1; fix in
  upstream commit \texttt{85961e4}, whose message does not mention it & Record; addendum
  ESP2 \\
Max-pool kernel ignores the lower bound of a fused activation when the channel count is 4
  modulo 8 & Kernel tests on shapes the models do not use & None; not reached by the files
  deployed at rows 1 to~4 & Reported with a fix and tests; merged upstream & Addendum ESP2 \\
Two further defects & Same tests & None; not reached by the files deployed at rows 1 to~4 &
  Reported privately; reproduced by Espressif; fixes due in 1.4.2 & Addendum ESP2 \\
Debug heap check inside inference, version 1.3.1 & Three heap walks per inference, at row~4
  only & Inside row~4's latency and energy of record: about \SI{0.77}{\milli\second} (1.4\%) in
  SRAM, \SI{1.0}{\milli\second} (1.1\%) in PSRAM & Footnoted wherever row~4 appears; rows 1
  to~3 unaffected & Addendum ESP2 \\
\midrule
\multicolumn{5}{@{}l}{\textit{Host scoring and the RP2040}} \\
TensorFlow Lite's default interpreter path does not reproduce the reference kernels & One file
  scored both ways: \SI{+1.0}{\pp}; 65 of 300 output vectors more than one step apart & Scored
  every INT8 accuracy; over 430 models, reference minus default averages \SI{-0.009}{\pp} (SD
  \SI{0.753}{\pp}); no registered contrast moves by more than \SI{0.16}{\pp}
  (\cref{sec:m-quant}) & Default path kept as the path of record & Record; addendum C5 \\
RP2040 outputs depart from the reference at the global average pool (MEAN) & Exact-match check
  failed at row~4; per-layer checksums & Class differs from the reference on 2 of 20 images at
  row~4 and 1 at row~3; RP2040 latencies come from an unverified path (\cref{sec:h-agree}) &
  Cause inside MEAN open; reference-kernel rebuild (P3) not run & Record; addendum P2, P4 \\
\bottomrule
\end{tabular}
\par\smallskip\raggedright\footnotesize\noindent\textit{Notes:} ``Record'' marks findings from
the measurements of record and ``addendum'' marks exploratory items by code. Supplementary Tables~S5
and~S6 give the log entries. The two privately reported defects are described no further until
Espressif's fix is public. Version numbers are those of the stack of record (\cref{tab:hwcond})
unless the row says otherwise.
\end{table}

\subsection{ESP32-S3 latency, memory and arena placement}\label{sec:h-esp}
\Cref{tab:ladder} gives what each deployed row costs on the ESP32-S3, all four rows timed on one
firmware lineage (\texttt{main\_T36.cc}). Latency runs from \SI{54134.0}{\micro\second} at row~4 to
\SI{858036.0}{\micro\second} at row~1, almost 16 times as long, and row~3 takes
\SI{152790.5}{\micro\second} in this lineage. No interquartile range is wider than
\SI{14.5}{\micro\second}, while firmware edits that left the model alone have moved a median by
several times that (\cref{sec:m-hw}), so the build limits these figures, not the timer. Per
multiply--accumulate the three MobileNetV2 rows cost 14.3 to \SI{16.3}{\nano\second}, highest at the
two PSRAM rows, and row~4, which has no depthwise layers, about \SI{7.1}{\nano\second}.
The row-4 figures in this subsection include ESP-NN 1.3.1's debug heap check (\cref{tab:toolchain}).

\begin{table}[htbp]
\centering
\caption{ESP32-S3 cost ladder, measured on one firmware lineage (\texttt{main\_T36.cc}).}
\label{tab:ladder}
\small
\setlength{\tabcolsep}{4pt}
\begin{tabular}{@{}l r r r l r r@{}}
\toprule
Row & \makecell[r]{Model size\\(\si{\byte})} & \makecell[r]{Arena used\\(\si{\byte})} &
\makecell[r]{Arena reserved\\(\si{\byte})} & Arena location &
\makecell[r]{Median latency\\(\si{\micro\second})} & \makecell[r]{IQR\\(\si{\micro\second})} \\
\midrule
1 & 2,715,632 & 576,044 & 1,048,576 & PSRAM & 858,036.0 & 14.5 \\
2 & 977,048 & 304,972 & 352,256 & PSRAM & 287,192.0 & 2.0 \\
3 & 627,464 & 224,476 & 281,600 & Internal SRAM & 152,790.5 & 11.0 \\
4 & 70,376 & 85,884 & 281,600 & Internal SRAM & 54,134.0 & 0.0 \\
\midrule
4, PSRAM control & 70,376 & 85,884 & 281,600 & PSRAM & 90,469.0 & 0.0 \\
\bottomrule
\end{tabular}
\par\smallskip\raggedright\footnotesize\noindent\textit{Notes:} ESP32-S3 at
\SI{240}{\mega\hertz} with ESP-NN 1.3.1 and the local patch (\cref{sec:m-hw}). Each model
is the INT8 arm-B file (B$'$ at row~4) for fold~0, seed~1, run on one input image.
Latency is the median and interquartile range (Tukey hinges) of 20 timed
\texttt{Invoke()} calls after one untimed warm-up. Arena used is the planner's reading;
arena reserved is the fixed block the firmware set aside. Location was read at start-up
from the arena pointer. The last line is row~4 with its arena moved into PSRAM at the
same reserved size and nothing else changed. Row~3's latency of record comes from another build and
is given in \cref{sec:h-esp}, not here.
At row~4, ESP-NN 1.3.1 runs a debug heap check three times per inference, about
\SI{0.77}{\milli\second} in SRAM (1.4\%) and \SI{1.0}{\milli\second} in PSRAM (1.1\%;
\cref{tab:toolchain}); row~4's latency and PSRAM factor include it, and without it the factor
would be about 1.676. Row~0 is not in this lineage; its addendum figures are in \cref{sec:h-esp}.
\end{table}

Arena use rises from \SI{85884}{\byte} at row~4 to \SI{576044}{\byte} at row~1. No static array
larger than \SI{290019}{\byte} would link, and that decides placement. Row~3 fits under it with
\SI{65543}{\byte} to spare, row~2 misses by \SI{14953}{\byte}, about 5\% of the limit, and row~1
needs roughly twice as much, so the step from SRAM to PSRAM falls between rows 2 and~3. Size and
placement change together up the ladder, and the two controls separate them only at rows 3 and~4.
With row~4's arena moved to PSRAM in the ladder's own lineage and nothing else changed, latency rose
to \SI{90469.0}{\micro\second}, 1.671 times its SRAM value. Row~3's control comes from an earlier
build in which the move to PSRAM also raised the reserved size, and there PSRAM added a factor of
1.351. The smaller arena paid the larger penalty, so I treat these as two measurements and do not
use either to estimate rows 1 and~2 in SRAM.

Row~3's latency of record comes from the \path{main_T24_final_PRE_ARENA_PROFILE.cc} harness, with
the same model, arena size and placement: \SI{152742.0}{\micro\second}, interquartile range
\SI{8.0}{\micro\second}. I use it whenever row~3's latency is quoted on its own and keep it out of
\cref{tab:ladder}. It differs from the ladder value through the build; the model is the same.

\paragraph{Addendum: the artefact of record, row~0 and placement \rc{exploratory}} Rebuilt on its
own harness, the row-3 artefact of record passed all three conditions set for it: arena, exact
outputs and median (item ESP1). Row~0 ran on the board for the first time (item ESP6), with a
\SI{1685852}{\byte} arena in PSRAM. Its outputs matched the reference on the twenty row-3
photographs, but 18 of the 20 reference vectors sit wholly on a limit, so the match shows little.
Its median latency in the repeat-capture harness (\texttt{main\_lat100\_rows.cc}) was
\SI{4057405}{\micro\second}. A \SI{304972}{\byte} internal allocation, row~2's arena, failed in all
three PSRAM builds (item ESP3): the largest free internal block was \SI{286720}{\byte},
\SI{18252}{\byte} short, so row~2 could not have run from internal SRAM as a heap block either.
Per-operator timing (item ESP5) put 82.6\% of row~4's PSRAM penalty on one convolution, which ran
5.1 times slower from PSRAM; without it the factor would be 1.135 instead of the 1.670 that the per-operator harness
measured. At rows 1 to~3
standard convolutions take 67 to 78\% of the time and depthwise ones 20 to 27\%.

\paragraph{Addendum: repeat latency, kernels and clocks \rc{exploratory}} The repeat captures (item
ESP7) timed every row again on \texttt{main\_lat100\_rows.cc}, 2,000 inferences over the twenty
row-3 photographs, a lineage of its own whose medians are in Supplementary Table~S4d. Within it
row~0 took 4.8 times as long as row~1, against 5.4 times the operations. Every arena sat where the
ladder put it, and every build matched its own reference on all twenty photographs. Three further
items measured ratios between builds (S4d). ESP-NN's optimised kernels were 4.7 to 6.8 times faster
than its own plain-C kernels (item ESP10), more at the SRAM rows; these ratios are not against a
build without ESP-NN, so they cannot be set beside Espressif's published speed-ups
(\cref{sec:rel-device}). The plain-C kernels need less arena, and at row~2 theirs,
\SI{231820}{\byte}, would fit under the static-array limit; I did not test that build. At rows 3
and~4, \SI{240}{\mega\hertz} was both the fastest CPU clock and the cheapest in gross energy (item
ESP11). A PSRAM clock of \SI{80}{\mega\hertz} booted ten cold boots out of ten with exact outputs
and cut latency by 11 to 19\% and gross energy by 8 to 13\% at the PSRAM rows (item ESP12).

\paragraph{Addendum: a newer ESP-NN stack \rc{exploratory}} Row~3 was also measured on ESP-NN 1.4.1
with esp-tflite-micro 1.4.1~\cite{esptflitemicro141}, unpatched (item ESP8). It matched the
reference on all twenty photographs, and its arena grew to \SI{245420}{\byte}, still in internal
SRAM. Its energy per inference was \SI{35.916}{\milli\joule} gross and \SI{15.282}{\milli\joule}
above idle, 7.5\% and 6.3\% below the repeat capture of the stack of record in the same warm-up
energy sketch. Its latency, timed on a harness of its own (\texttt{main\_lat100.cc}), a lineage no
other figure shares, had a median of \SI{138954.0}{\micro\second}. Both components changed, so this
compares two stacks, not two ESP-NN versions.

\subsection{RP2040 flash-capacity limit}\label{sec:h-rp}
\Cref{tab:rp2040} gives the outcome for each row on the RP2040 at \SI{133}{\mega\hertz}. Row~1 did
not link. Rows 2 to~4 linked, allocated their arenas and ran all 20 timed inferences, and none
failed on RAM. Row~2 ran, against my prediction that it would fail. An arena is not a property of
the model alone: for row~2 the RP2040's planner took \SI{232412}{\byte} against \SI{304972}{\byte}
on the ESP32-S3, and on the ESP32-S3 itself row~3 needs \SI{204748}{\byte} with plain-C kernels and
\SI{224476}{\byte} with optimised ones.

\begin{table}[htbp]
\centering
\caption{RP2040 outcome and cost for each row (Raspberry Pi Pico~H at
\SI{133}{\mega\hertz}). Sizes are in bytes.}
\label{tab:rp2040}
\small
\setlength{\tabcolsep}{4pt}
\begin{tabular}{@{}l r l r r r r r r@{}}
\toprule
Row & \makecell[r]{Model\\size} & Outcome &
\makecell[r]{Flash\\used} & \makecell[r]{RAM\\used} &
\makecell[r]{Arena\\granted} & \makecell[r]{Arena\\used} &
\makecell[r]{Median\\latency\\(\si{\micro\second})} & \makecell[r]{IQR\\(\si{\micro\second})} \\
\midrule
1 & 2,715,632 & Flash overflow & -- & -- & -- & -- & -- & -- \\
2 & 977,048 & Runs & 1,286,948 & 250,248 & 235,000 & 232,412 & 2,659,899.0 & 184.5 \\
3 & 627,464 & Runs & 937,372 & 224,144 & 208,896 & 205,332 & 1,795,947.0 & 133.5 \\
4 & 70,376 & Runs & 359,548 & 146,320 & 131,072 & 49,348 & 1,096,854.5 & 201.5 \\
\bottomrule
\end{tabular}
\par\smallskip\raggedright\footnotesize\noindent\textit{Notes:} Built in Arduino IDE~2
with arduino-pico 5.6.0 (GCC 14.3) and MicroTFLite 1.0.4, using the all-operators resolver
and CMSIS-NN kernels with no acceleration path. The models are the INT8 arm-B files (B$'$
at row~4) for fold~0, seed~1, as in \cref{tab:ladder}, each run on one input image. Flash
used and RAM used are the IDE's sketch and global-variable totals, out of a
\SI{2093056}{\byte} flash region and \SI{262144}{\byte} of RAM. Arena granted is the fixed array
compiled
into each build; arena used is the planner's reading at that size and is not a minimum.
Latency is the median and interquartile range (Tukey hinges) of 20 timed
\texttt{ModelRunInference()} calls. The outputs are not bit-exact with the ESP32-S3's
(\cref{sec:h-agree}), so the kernel path behind these latencies is unverified, and they can
be compared with ESP32-S3 figures only to order of magnitude. Row~0 was not built for the
RP2040.
\end{table}

Row~1 failed on flash: the \SI{2093056}{\byte} region overflowed by \SI{949388}{\byte}. The model
file alone, \SI{2715632}{\byte}, is \SI{622576}{\byte} larger than the whole region, so no trimming
of code could have saved it, and the build stopped before RAM came into it. RAM was never the
binding limit. Every arena I granted was below the largest that would link (\SI{244840}{\byte}), and
row~2 ran with its static data filling 95.5\% of RAM. The arena-used figures are planner readings,
not minima: rebuilt with exactly that much arena, rows 2 and~3 failed at start-up 4 bytes short, so
their minima lie between the reading plus 4 bytes and the grant that worked. The addendum traced the
4-byte change in fixed overhead to padding after the arena when its size is 4 more than a multiple
of 8 (item P5) \rc{exploratory}. That padding is not shown to be the allocator's missing 4 bytes.

Latency runs from \SI{1096854.5}{\micro\second} at row~4 to \SI{2659899.0}{\micro\second} at row~2,
with no interquartile range wider than 0.02\% of its median. Against the ESP32-S3 ladder
(\cref{tab:ladder}) the RP2040 is 20.3, 11.8 and 9.3 times slower at rows 4, 3 and~2. The row-4 ratio 
uses an ESP32-S3 latency that includes the debug heap check (\cref{tab:toolchain}). Row~3's
latency of record gives 11.8 as well. Each ratio folds together the instruction set, the clock, the
kernel library and the TFLM snapshot, so it supports one claim only: the RP2040 is about an order of
magnitude slower on these models. The kernel path behind it also does not reproduce the reference
arithmetic (\cref{sec:h-agree}). Row~4's latency moved by up to 0.38\% across three builds, against
0.03\% at rows 2 and~3, so I quote it with its build. With interrupts left on, as in every figure
here, servicing them took 0.74, 0.68 and 2.52\% of inference time at rows 2, 3 and~4; I did not
separate the sources.

\paragraph{Addendum: repeat latency and clock \rc{exploratory}} Reflashed, rows 2 to~4 reproduced
their \SI{133}{\mega\hertz} output fingerprints exactly (first part of item P1; the meter check is in 
\cref{sec:m-design}). The repeat captures (item P6) timed rows 2 to~4 again for arms B and~D, 2,000 inferences 
per build on a harness of their own (\texttt{P6\_latency.ino}; medians in Supplementary Table~S4d). Within 
that lineage the arms differ by at most \SI{25}{\micro\second} and use the same arena. At \SI{200}{\mega\hertz}, 
rows 3 and~4 ran faster in proportion to the clock, to within 0.1\% (item P7).

\subsection{Energy per inference}\label{sec:h-energy}
\Cref{tab:energy} gives the energy of one inference for every deployed row on both boards, each the
mean of the 30 windows in one capture of the arm-B file (B$'$ at row~4), idle-subtracted first with
the gross figure beside it. Energy on the RP2040 is a post-registration addition.
Every ratio in this subsection that involves ESP32-S3 row~4 includes ESP-NN 1.3.1's debug heap check
(\cref{tab:toolchain}).

\begin{table}[htbp]
\centering
\caption{Energy per inference on both boards: mean and standard deviation over the 30
windows of one capture.}
\label{tab:energy}
\footnotesize
\setlength{\tabcolsep}{3pt}
\begin{tabular}{@{}l l l r r r r@{}}
\toprule
Board & Row & Arena & \makecell[r]{Idle-subtracted\\(\si{\milli\joule})} &
\makecell[r]{Gross\\(\si{\milli\joule})} & \makecell[r]{Idle share\\of gross (\%)} &
\makecell[r]{Power above\\idle (\si{\milli\watt})} \\
\midrule
ESP32-S3 & 1 & PSRAM & 74.025 (0.435) & 201.209 (0.436) & 63.2 & 86.5 \\
         & 2 & PSRAM & 27.587 (0.087) & 69.969 (0.086) & 60.6 & 95.7 \\
         & 3 & SRAM & 16.322 (0.058) & 38.641 (0.055) & 57.8 & 106.8 \\
         & 4 & SRAM & 6.665 (0.039) & 14.582 (0.045) & 54.3 & 123.9 \\
         & 4, PSRAM control & PSRAM & 9.058 (0.037) & 22.374 (0.034) & 59.5 & 100.4 \\
\midrule
RP2040\textsuperscript{a} & 2 & SRAM & 41.211 (0.201) & 243.156 (0.205) & 83.1 & 15.4 \\
         & 3 & SRAM & 27.292 (0.194) & 162.897 (0.193) & 83.2 & 15.1 \\
         & 4 & SRAM & 19.299 (0.083) & 101.179 (0.083) & 80.9 & 17.6 \\
\bottomrule
\end{tabular}
\par\smallskip\raggedright\footnotesize\noindent\textit{Notes:} ESP32-S3 at
\SI{240}{\mega\hertz} and RP2040 (Pico~H) at \SI{133}{\mega\hertz}, both supplied at
\SI{3.3}{\volt} into the 3V3 pin (\cref{tab:hwcond}). Each figure comes from one capture
of the INT8 arm-B file (B$'$ at row~4) for fold~0, seed~1, on one input image.
Gross and idle-subtracted energy follow \cref{eq:egross,eq:einc}; idle-subtracted energy
is energy above this firmware's own idle, not above a sleeping
chip. Power above idle is idle-subtracted energy divided by the mean window length. Row~1
does not fit on the RP2040, and row~0's addendum figures are in \cref{sec:h-energy}. Row~4 on the
ESP32-S3 includes ESP-NN 1.3.1's debug heap check (\cref{tab:ladder}).
\textsuperscript{a}Post-registration addition, captured with the on-board regulator shut
down. The same rows supplied at \SI{5.0}{\volt} through the regulator are in
Supplementary Table~S4a.
\end{table}

\paragraph{ESP32-S3} Idle-subtracted energy rose from \SI{6.665}{\milli\joule} at row~4 to
\SI{74.025}{\milli\joule} at row~1, with \SI{16.322}{\milli\joule} at row~3; gross energy ran from
\SI{14.582}{\milli\joule} to \SI{201.209}{\milli\joule}, with \SI{38.641}{\milli\joule} at row~3.
Idle draw is 54.3 to 63.2\% of each gross figure: more than half of what the meter records during an
inference would have been drawn anyway. Within a capture the windows agree to a standard deviation
of at most 0.59\% of the mean. Row~3 came from a different session, and the other session's two
row-3 captures came out 0.37 and 0.15\% above it in idle-subtracted energy and 0.54 and 0.70\% in
gross, so I do not compare energies across sessions more finely than that. Energy grew more slowly
than the window: from row~4 to row~1 the window lengthened 15.9 times while idle-subtracted energy
rose 11.1 times and gross energy 13.8 times, and power above idle fell from \SI{123.9}{\milli\watt}
to \SI{86.5}{\milli\watt}.

The row-4 control separates placement from size for one model. With its arena moved to PSRAM the
window lengthened by a factor of 1.678 and idle-subtracted energy rose by a factor of 1.359, or
\SI{2.393}{\milli\joule} per inference, while gross energy rose by a factor of 1.534. For this model
PSRAM cost proportionally less in energy than in time, and with no second energy point I do not
carry the factor to rows 1 and~2. Row~1's capture idled at \SI{45.04}{\milli\ampere}, 0.30 to
\SI{0.50}{\milli\ampere} above the rest of its session for reasons I could not establish; if its
true idle matched the rest of its session, its idle-subtracted figure would be 1.2 to 1.9\% higher.

\paragraph{RP2040} Supplied at the 3V3 pin with the regulator shut down (\cref{sec:m-hw}), the
RP2040 used \SI{19.299}{\milli\joule}, \SI{27.292}{\milli\joule} and \SI{41.211}{\milli\joule} above
idle at rows 4, 3 and~2, and 101.179, 162.897 and \SI{243.156}{\milli\joule} gross, of which idle is
81 to 83\%. That is 2.90, 1.67 and 1.49 times the ESP32-S3's idle-subtracted energy and 6.94, 4.22
and 3.48 times its gross energy. Each inference takes about ten times longer, but power above idle
is only 15.1 to \SI{17.6}{\milli\watt}, six to seven times lower than the ESP32-S3's, while the idle
current of 22.6 to \SI{22.9}{\milli\ampere} flows for the whole of the longer window. These figures
are more fragile. Inference raised the current by only 4.6 to \SI{5.3}{\milli\ampere}, so an error
of \SI{1}{\milli\ampere} in the idle estimate would move the idle-subtracted figure by 19 to 22\%,
and the RP2040 captures followed \SI{1800}{\second} of warm-up where the ESP32-S3 captures of record
followed none, which every ratio between the boards carries. Through the regulator in PWM mode
instead, gross energy was about 1.5 times as large but idle-subtracted energy only 1.1 to 1.6\%
higher (Supplementary Table~S4a), some reassurance for the subtraction.

\begin{figure}[htbp]
\centering
\includegraphics[width=0.6\linewidth]{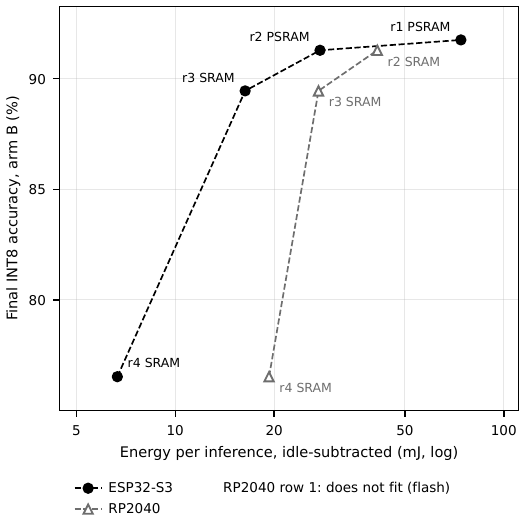}
\caption{INT8 accuracy of arm~B against idle-subtracted energy per inference, on the
ESP32-S3 (filled circles) and the RP2040 (open triangles); labels give the row and arena
placement, and the dashed lines join each board's Pareto frontier. Energy comes from one
capture of one deployed file per row; accuracy is the ten-fold mean on the host's default
path (three seeds at rows 3 and~4). RP2040 energy is a post-registration addition, and row~1
does not fit in its flash. Row~4 on the ESP32-S3 includes ESP-NN 1.3.1's debug heap check
(\cref{tab:toolchain}). The horizontal axis is logarithmic.}
\label{fig:pareto}
\end{figure}

\paragraph{Accuracy against energy} \Cref{fig:pareto} sets each row's energy against arm~B's INT8
accuracy. On both boards every row sits on that board's Pareto frontier. On the ESP32-S3, row~4 to
row~3 added 12.92 points for 2.45 times the energy, row~3 to row~2 added 1.84 points for 1.69 times,
and row~2 to row~1 added 0.47 points for 2.68 times, less than the run-to-run standard deviation of
1.33 points for arm~B at row~3 (\cref{tab:spread}). The arm does not enter this picture
(\cref{sec:invariance}).

\paragraph{Addendum: repeat captures \rc{exploratory}} Every deployed row was captured again on both
boards after \SI{1800}{\second} of warm-up (items ESP7 and P6), with arm~D beside arm~B at rows 1, 2
and~4 on the ESP32-S3 and rows 2 to~4 on the RP2040 (D$'$ beside B$'$ at row~4), and row~0 on the ESP32-S3 for the first time:
\SI{957.13}{\milli\joule} gross and \SI{354.79}{\milli\joule} above idle. All 24 ESP32-S3 and 18
RP2040 captures were valid with no dropped sample, and captures of one build agreed within 0.15 to
0.91\% gross on the ESP32-S3 and within 0.36\% on the RP2040. Against the values of record, arm~B
moved by at most 0.73\% gross and 0.49\% idle-subtracted on the ESP32-S3, despite a different
warm-up and session, and by at most 0.77\% and 0.36\% on the RP2040. Supplementary Tables~S4a and~S4d
give every figure. At rows 3 and~4, raising the RP2040's clock to \SI{200}{\mega\hertz} (item P7) lowered
gross energy by 2.4 and 3.0\% but raised idle-subtracted energy by 9.2 and 6.8\%, with idle current
43\% higher.

\section{Arm invariance of deployment cost}\label{sec:invariance}
The registration says that ``the hardware metrics are architecture-determined rather than
arm-determined''~\cite{yadav2026osf}, Wilson states the same (\cref{sec:rel-device}), and the ladder
in \cref{sec:hardware} relies on it, because every file deployed there is an arm-B file. I checked
it at row~3 on the ESP32-S3, building the arm-A and arm-D models for fold~0, seed~1, against the
criteria of \cref{sec:m-hw}; arm~C was not built. \Cref{tab:invariance} gives the results. The three
files are far from near-copies: on fold~0 arm~A's INT8 file scores 60.3\%, arm~B's 87.3\% and
arm~D's 88.3\%.

\paragraph{Memory and latency} The arena used \SI{224476}{\byte} of internal SRAM in all three
builds, which meets that criterion exactly. The model files were not all the same size, and that
criterion was missed: arm~D's file is \SI{627472}{\byte} against \SI{627464}{\byte} for A and~B, and
the 8 bytes are tensor names, with every weight buffer at the same offset (Supplementary Note~D). On
the chip the ESP-IDF size report was the same in every section for arms A and~D, and in the energy
firmware all three images are the same size, \SI{963312}{\byte}. Arm~A's median latency was
\SI{152790.5}{\micro\second}, the same as arm~B's in \cref{tab:ladder}, and arm~D's
\SI{152818.5}{\micro\second}. The largest difference, \SI{28}{\micro\second} or 0.018\%, is inside
the \SI{\pm 100}{\micro\second} bound, and the criterion was met. The offset is real, since arm~A's
20 timings and arm~D's do not overlap, but the bound was set by link layout and the test shows
invariance at that resolution only.

\paragraph{Outputs} Each arm's six outputs matched its own reference exactly, which shows the right
weights were linked. Arm~A predicts bathroom for the living-room image, and the board reproduced
that wrong answer with none of its outputs on a limit.

\paragraph{Energy} Captured in the order A, B, D, A, gross energy per inference was
\SI{38.690}{\milli\joule}, \SI{38.641}{\milli\joule}, \SI{38.651}{\milli\joule} and
\SI{38.914}{\milli\joule}. The statistic of \cref{eq:armR} was $R = +0.100\%$ (95\% interval
$+0.019\%$ to $+0.181\%$), and the two arm-A captures differed by 0.58\% of arm~B's figure. Both
guards held and $|R|$ was far under 1\%. The test passed. The interval reflects only the spread
between windows, though, and two captures of the same arm differed by about six times $R$; the
second arm-A capture would give $R = +0.68\%$, still inside the bound. Idle-subtracted, the first
arm-A capture was 0.33\% above arm~D's, against 0.16\% between the two arm-A captures.

\paragraph{Addendum: the 8 bytes and the \SI{28}{\micro\second} \rc{exploratory}} Converted in a
fresh process, arm~B came out at \SI{627464}{\byte}, and converted after arm~B had been built, arm~D
came out at \SI{627472}{\byte}. Both are byte for byte the registered files. Reversing the order
gave arm~D at \SI{627464}{\byte} and arm~B at \SI{627472}{\byte}, so the length comes from Keras
naming state at conversion and not from the arm (item C3). The longer arm-B file, timed on the ladder
harness, came out \SI{28.7}{\micro\second} above arm~A, against \SI{28.0}{\micro\second} for arm~D
(item ESP9). The offset follows the file and not the arm. The linker maps show 115 sections moved by
8 bytes, and where the shift is absorbed was not located.

\paragraph{Addendum: arms B and~D at the other rows \rc{exploratory}} The repeat captures built
arm~D beside arm~B at rows 1, 2 and~4 on the ESP32-S3 and rows 2 to~4 on the RP2040, D$'$ beside
B$'$ at row~4 (items ESP7 and P6). On both boards the arena was the same for both arms at every row, and each arm matched its own
reference on all twenty photographs. On the ESP32-S3 the model files were the same size at rows 2
and~4; at row~1 arm~D's was 8 bytes shorter, again tensor names. On the RP2040 the arms differed by
at most 0.53\% in gross energy and 0.12\% idle-subtracted, and row~3 showed no file-length offset.
On the ESP32-S3, arm~D's gross energy was 0.82\% above arm~B's at row~1 and 0.47\% at row~2, inside
the 1\% bound, and at row~4 D$'$ was 1.44\% above B$'$, outside it (0.38\% idle-subtracted).
Idle current was 2.3\% higher for D$'$, and the arms were captured in separate, unbracketed sessions, so drift could
explain the gap, and the pair is unresolved. Latency on the ESP32-S3, in
\texttt{main\_lat100\_rows.cc}, is less tidy: arm~D's median was \SI{2405}{\micro\second} (0.28\%)
above arm~B's at row~1, \SI{1088}{\micro\second} (0.38\%) below at row~2 and
\SI{12.5}{\micro\second} below at row~4 (D$'$ against B$'$). No latency bound was declared for these rows, the row-1 and
row-2 gaps are many times the interquartile range of about \SI{58}{\micro\second}, file length
cannot explain the row-2 gap, and I did not find the cause.

\paragraph{Reading the check with the addendum} The row-3 result is unchanged, and the
\SI{28}{\micro\second} offset there comes from the file. Memory and energy follow the model and the
build, not the arm, on both boards and at every row checked, within 1\% in energy except the
ESP32-S3 pair at row~4, which I cannot resolve. On the ESP32-S3 the arms' latencies differ by under
0.4\% at rows 1 and~2 for reasons I did not find, so I do not claim that latency is independent of
the arm there. For deployment, the choice between arms B and~D still comes down to accuracy.

\begin{table}[htbp]
\centering
\caption{Deployment cost of the row-3 model for arms A, B and~D on the ESP32-S3, against
the criteria fixed before the device runs.}
\label{tab:invariance}
\footnotesize
\setlength{\tabcolsep}{3pt}
\begin{tabular}{@{}l r r r l l@{}}
\toprule
Check & Arm A & Arm B & Arm D & Criterion & Outcome \\
\midrule
Arena used (\si{\byte}), location & 224,476, SRAM & 224,476, SRAM & 224,476, SRAM &
  Identical & Met, exact \\
Model file (\si{\byte}) & 627,464 & 627,464 & 627,472 & Same size &
  Missed\textsuperscript{a} \\
Median latency (\si{\micro\second}) & 152,790.5 & 152,790.5\textsuperscript{c} & 152,818.5 &
  Within \SI{\pm 100}{\micro\second} & Met \\
IQR (\si{\micro\second}) & 12.5 & 11.0\textsuperscript{c} & 13.0 & -- & Reported \\
Outputs equal to own reference & 6/6 & 6/6 & 6/6 & Exact & Met \\
Outputs on an INT8 limit & 0/6 & 5/6 & 3/6 & -- & Reported \\
Predicted class & Bathroom & Living room & Living room & -- & Reported \\
Gross energy (\si{\milli\joule}) & 38.690; 38.914\textsuperscript{b} & 38.641 & 38.651 &
  $|R| \leq 1\%$ & Passed \\
\bottomrule
\end{tabular}
\par\smallskip\raggedright\footnotesize\noindent\textit{Notes:} MobileNetV2
($\alpha = 0.35$, \SI{96}{px}), fold~0, seed~1, full real data, INT8, one photograph of a
living room, ESP32-S3 at \SI{240}{\mega\hertz}; protocols as in \cref{tab:ladder,tab:energy}.
Energy captures ran in the order A, B, D, A, with $R = +0.100\%$ (\cref{eq:armR}).
\textsuperscript{a}The 8~bytes are tensor names, with every weight buffer at the same
offset in all three files (\cref{sec:invariance}; Supplementary Note~D). \textsuperscript{b}First and second arm-A captures.
\textsuperscript{c}From the ladder build (\texttt{main\_T36.cc}, \cref{tab:ladder}), made earlier the same
day as the arm-A and arm-D builds; arm~B was not rebuilt for this check.
\end{table}

\section{Discussion}\label{sec:discussion}
Does synthetic pre-training pay off on a microcontroller? With all the real data and the frozen
recipe, no. The primary interval stops below the planning value recomputed from Bird et al., the
only full-data difference the registered tests resolved is a deficit of about 1.3 images per fold
at row~4, and the trend regression found no detectable dependence on model size; the addendum's
three-replicate refit leans positive but does not survive adjustment, and I leave that open. The
anchor, with a quarter of the photographs, failed too, though over four synthetic draws its deficit
was about \SI{0.6}{\pp}, with an interval reaching just above zero \rc{exploratory}. Rows 0 to~3
all sit above the size at which Muralidharan et al.\ found pre-training stopped paying
reliably~\cite{muralidharan2026}, which makes capacity alone an unlikely explanation for the nulls,
although their contrast is arm~B against arm~A and not D against~B. The full-data result also
belongs to the recipe: with heavier augmentation the row-3 null became a deficit and the row-4
deficit disappeared (\cref{sec:explore}).

The effect is not one fixed number, though. With a tenth of the real photographs the row-4 CNN
gained on every fold, by 6 to 19 images on seed~1, and the gain replicated; neither the weak
Imagenette start alone nor the synthetic labels explain it, and the CNN's \SI{48}{px} input has not
been separated from its size (\cref{sec:fraction}). So the answer to the title is narrow: not at full real data
under this recipe, and with scarce real data a replicated gain for the smallest model only, for
reasons this study has not identified. The next tests are about the CNN: running it at \SI{96}{px},
or a MobileNetV2 at \SI{48}{px}, would separate input size from model size, each with its test fixed
before the runs.

Quantisation did not change the picture, although the design was built to look for it. INT8
conversion cost the row-3 arms 2.1 to 2.4 points, more than the gain the study was planned around,
yet the difference-in-differences was null, and both row-4 results whose intervals clear zero have
the same sign in float32. At row~3 there was no float32 gain for INT8 to take away, so that null is a bound
on movement, not evidence that a benefit survives.

On the device the arms cost the same where I could check, with two small exceptions on the
ESP32-S3 that I could not explain: at row~4 arm~D$'$ drew 1.44\% more gross energy than B$'$, outside
the 1\% bound, and at rows 1 and~2 the arms' latencies differ by under 0.4\% \rc{exploratory;
\cref{sec:invariance}}. The synthetic stage is paid for offline, once to build the
rendered set and then in a few GPU-seconds per run, and showed no run-time cost that matters. Model
and build matter far more: model size sets latency and where the ESP32-S3's arena can sit, flash
capacity rather than RAM binds the RP2040, and kernel choice, clock and even harness edits move
latency measurably, so anyone repeating these measurements should take every baseline from the same
build.

The checks behind those figures carry three lessons for deploying models like these. A match on the
final output is weak evidence when the output saturates, and only the pre-softmax probe and the
per-layer checksums tested the arithmetic underneath. A board can disagree with the reference
without anything going visibly wrong: the RP2040's outputs summed to one and mostly picked the
reference's class, yet the build departs at the global average pool, and the host path that scored
every accuracy is not the reference arithmetic either. And the vendor kernels themselves can be
wrong: the overrun that crashed row~3 was removed upstream in a commit that does not mention it, and
tests on shapes my models do not use found three more defects. None of this moved a registered
result, but a quantised model's accuracy on a board is something to check against the reference, not
read off the host.

\paragraph{Hypothesis} One explanation fits this pattern, and nothing in this study tests it. The
rendered set is less varied than its 5,928 images suggest, coming from 23 camera placements with
neighbouring frames 5\textdegree{} apart (\cref{sec:m-data}), so Bird et al.'s 98\% accuracy on it
may say more about how alike neighbouring frames are than about what the network learned. A stage
like that would add little to features already learned from real photographs, and might pull them
slightly off course, which would fit the deficits at the anchor and at row~4; on a weak start with
few real photographs even low-variety scene statistics could help, which would fit the row-4 gain at
10\%. The shuffled-label control points the same way, at the images rather than the classes. Work
outside microcontrollers suggests variety matters~\cite{tobin2017,baradad2021}, though not simply:
Mishra et al.\ found that with every variation switched on, fine-tuned accuracy fell below training
from scratch (62.96\% against 64.85\%)~\cite{mishra2022}. Whether the number of distinct scenes, at
a fixed number of rendered images, determines whether a synthetic stage pays off at microcontroller
scale is a question for a second study.

\section{Limitations}\label{sec:limits}
This is a case study: one dataset of six scene classes~\cite{bird2020}, one renderer, one recipe and
one board of each kind. I make no claim about other tasks, newer image generators, or Cortex-M parts
with DSP or vector extensions, and with no third-party results on the dataset the absolute
accuracies have no external baseline. Some of the web photographs may also be in ImageNet, which I
had no practical way to check. Imagenette is drawn from ImageNet, so at every row any overlap
reaches arms B and~D alike, though I cannot show that it cancels. The folds are stratified by class,
not grouped by scene, and a 30-photograph spot check is not a measurement. A looser duplicate screen
found 18 pairs split across folds that I did not inspect; dropping them moved neither INT8
confirmatory contrast by more than \SI{0.03}{\pp}, but no float32 test predictions were saved, so the
anchor and the difference-in-differences could not be rechecked (item C15, \cref{sec:m-data})
\rc{exploratory}.

The real-data recipe was frozen before the grid, and the addendum showed that a second recipe moves
the arms differently (\cref{sec:explore}), so the variance that comes from choosing
hyperparameters~\cite{bouthillier2021} is not sampled. Quantisation was post-training only, with
plain calibration and no quantisation-aware training (\cref{sec:m-quant}), so the
difference-in-differences describes this pipeline.

Three choices used data that later served as test folds. The 42-epoch budget came from a pilot on
fold~0, but it used arm~A and one budget applies to every arm, so it cannot favour either side of a
paired comparison. The synthetic-stage length and freezing were tuned on arm~D on folds 0 to~2 with
no matching budget for arm~B; both rules ended in ties and kept the cheaper setting, and any residual
optimism would favour arm~D. Leaving those folds out, the row-3 difference stayed null, but over
seven folds the INT8 upper bound, \num{+1.710}, no longer excludes the \SI{+1.6}{\pp} planning value
(item C14, \cref{sec:explore}) \rc{exploratory}, so the registered bound is a narrow one.

The Nadeau and Bengio correction~\cite{nadeau2003} allows only approximately for the eight of
nine training folds that any two runs share~\cite{dietterich1998}, so every interval's coverage is
approximate. The folds replicate the real-data stage but not the synthetic one, and the registered
anchor rests on one synthetic draw. Its fail stands as measured. Over four draws the deficit was
about \SI{0.6}{\pp}, with an interval reaching \SI{+0.031}{\pp} (item C7, \cref{sec:explore})
\rc{exploratory}, so neither result supports a claim that the stage reliably costs the largest model
a point. The decision to stay at three seeds rested on a five-seed pilot whose standard deviation
has a relative error of about a third, although the grid's $\sigma_d$ came out close to it. The
exploratory sections looked at many contrasts without adjustment, and the addendum's screen kept one
of 16, so any single exploratory result is a lead, not a finding.

Every hardware figure of record comes from one deployed file per row (fold~0, seed~1) on one image,
and row~4's ESP32-S3 latency and energy include ESP-NN 1.3.1's debug heap check
(\cref{tab:toolchain}). The RP2040's latencies come from a kernel path that does not reproduce the
reference arithmetic, and what inside its global average pool causes the departure was not tested
(\cref{sec:h-agree}). The energy procedure is not MLPerf Tiny's~\cite{mlperftiny}, and the idle
current includes board components I did not separate. No capture measured the supply voltage, so
every absolute millijoule assumes the nominal \SI{3.3}{\volt}, and without the meter check of item P1
drift in the meter cannot be told from drift in the RP2040 (\cref{sec:m-hw}). The ESP32-S3 captures
of record had no warm-up and the RP2040's had \SI{1800}{\second}, which every ratio between the
boards carries. Arm invariance was tested against declared criteria only at row~3 on the ESP32-S3,
and arm~C was never built. At the other rows the addendum declared no latency bound, and the row-4
energy pair on the ESP32-S3 is unresolved (\cref{sec:invariance}).

\section{Conclusion}\label{sec:conclusion}
I asked whether a synthetic pre-training stage, run between the ImageNet weights and the real
photographs, helps a classifier small enough for a microcontroller. At full real data and under the
recipe used, it did not: the main model's INT8 interval excludes the \SI{+1.6}{\pp} planning value,
the smallest model showed a small deficit, and quantisation did not detectably change the difference
between the arms. With a tenth of the real photographs the smallest model gained on every fold, a
gain that replicated in the exploratory addendum and is explained neither by its weak starting
weights alone nor by the synthetic labels; its input resolution has not been separated from its
size, and its cause is open. On the ESP32-S3 the stage moved neither latency nor energy beyond the
bounds set before the runs at row~3, and the small arm differences found at other rows are
unexplained, while model size, memory placement, the RP2040's flash-capacity limit and the kernel
build decided the cost. The ESP32-S3 computed exactly what the reference kernels compute once a
vendor kernel was patched; the RP2040 and the host interpreter did not.

What is left is to separate the small CNN's size from its input resolution in a test fixed in
advance, to vary the number of distinct rendered scenes at a fixed image count, to find what in
the RP2040's global average pool departs from the reference, to repeat the ESP32-S3 row-4 energy
pair in one bracketed session, and to add quantisation-aware training and a Cortex-M33 part such
as the RP2350 to the comparison.

\section*{CRediT authorship contribution statement}
\textbf{Achintya Yadav:} Conceptualization, Methodology, Software, Validation, Formal analysis,
Investigation, Resources, Data curation, Writing -- original draft, Writing -- review \& editing,
Visualization, Project administration.

\section*{Declaration of competing interest}
The author has no known competing financial interests or personal relationships that could have
appeared to influence the work reported in this paper.

\section*{Funding}
This research did not receive any specific grant from funding agencies in the public, commercial, or
not-for-profit sectors.

\section*{Declaration of generative AI and AI-assisted technologies in the manuscript preparation
process}
During the preparation of this work the author used Anthropic's Claude Sonnet 4.8 and~5 to correct
language and grammar, improve readability and check formatting. After using this tool, the author
reviewed and edited the content as needed and takes full responsibility for the content of the
published article.

\section*{Data availability}

Data will be made available on request. This covers the code, the per-run results files of the
registered study and the addendum, the checkpoint weights, the firmware and the raw energy captures,
from which every reported value can be re-derived.

The code and hash files as filed with the addendum on 5 October 2026, before any of its items ran,
are attached to its registration~\cite{yadav2026osfaddendum}.

I do not share the images, because they are not mine to share. They come from the dataset of Bird et
al.~\cite{bird2020data}, which is on Kaggle
(\path{birdy654/environment-recognition-simulation-to-reality}), and \texttt{folds.csv} gives the
fold of every image that remained after deduplication, so the splits can be rebuilt from the
original download. Imagenette was obtained through TensorFlow Datasets (160px-v2), and the ImageNet
pre-trained MobileNetV2 weights come from Keras; the weights manifest records their checksums.

The project log is not shared; Supplementary Tables~S5 and~S6 give the departures and the addendum
items in full. The fix and regression tests for the public max-pool defect are in the upstream pull
request~\cite{espnnpr42}.

\section*{Supplementary material}

Six tables, three figures and a PDF of supplementary notes are uploaded with the manuscript and
published as received. The notes PDF opens with a full description of every table and figure. The
tables are UTF-8 CSV files. Each marks its exploratory rows, and all RP2040 energy in Table~S4 is
a post-registration addition. Differences are D$-$B (D$'-$B$'$ at row~4) in percentage points,
with the corrected 95\% intervals of \cref{sec:m-stats}.

\paragraph{Supplementary Table S1: per-fold accuracy} Fold-by-fold test accuracy of the registered runs,
for every row, arm, precision, seed and data fraction, with the anchor's second synthetic draw and
the calibration bridge. The addendum's runs are not listed.

\paragraph{Supplementary Table S2: contrasts between arms} All 60 pairwise contrasts at full data,
the difference-in-differences at every row, per-seed contrasts, the seed-1 inputs to the trend
regression and the row-0 follow-ups of \cref{sec:explore}, then the \rc{exploratory} addendum
results (C7 to C11), each beside the registered value it bears on.

\paragraph{Supplementary Table S3: re-analysis of the source study} The twelve rows of Table~II in
Bird et al.~\cite{bird2020} with printed and recomputed differences and the working range.

\paragraph{Supplementary Table S4: hardware and energy records} S4a every energy capture with its
hash where one is on record, including the failed and alternative RP2040 sessions and the 87 \rc{exploratory} addendum
captures; S4b the post hoc sanity checks; S4c the full build and measurement conditions; S4d the
addendum's repeat latency (each in its own lineage), build and clock ratios and repeat energy. Row-4
ESP32-S3 values from builds with the optimised kernels carry the ESP-NN 1.3.1 heap-check footnote,
and RP2040 energy is a post-registration addition.

\paragraph{Supplementary Table S5: departures from the registered plan} Every logged departure that
bears on a reported result, with its entry number, the corrections to values first filed in the log,
the departures inside the addendum's items and the breakdown of the run budget.

\paragraph{Supplementary Table S6: addendum items} One row per addendum item, keyed as in
\cref{sec:m-design}: claim, design, count, reading rule, family, status, outcome, log entries and departures from the declaration,
with C10's row-0 arm and P3 listed as not run and the P1 meter check as not done.

\paragraph{Supplementary Notes A to E} A, the dataset; B, recipe decisions; C, the
representative-set rule; D, hardware and arm-invariance detail; E, exploratory follow-ups.

\paragraph{Supplementary Figures S1 to S3} The seed-1 trend inputs against model size, the
synthetic-epoch sweep, and the \rc{exploratory} three-replicate trend with raw and BH-adjusted $p$.

\section*{Acknowledgements}
I thank Bird et al.~\cite{bird2020} for releasing the dataset on which this study rests.

\bibliographystyle{elsarticle-num}
\bibliography{refs}

\begin{thebibliography}{10}
\expandafter\ifx\csname url\endcsname\relax
  \def\url#1{\texttt{#1}}\fi
\expandafter\ifx\csname urlprefix\endcsname\relax\def\urlprefix{URL }\fi
\expandafter\ifx\csname href\endcsname\relax
  \def\href#1#2{#2} \def\path#1{#1}\fi

\bibitem{deng2009}
J.~Deng, W.~Dong, R.~Socher, L.-J. Li, K.~Li, L.~Fei-Fei,
  \href{https://doi.org/10.1109/CVPR.2009.5206848}{{ImageNet}: A large-scale
  hierarchical image database}, in: 2009 IEEE Conference on Computer Vision and
  Pattern Recognition (CVPR), 2009, pp. 248--255.
\newblock \href {https://doi.org/10.1109/CVPR.2009.5206848}
  {\path{doi:10.1109/CVPR.2009.5206848}}.
\newline\urlprefix\url{https://doi.org/10.1109/CVPR.2009.5206848}

\bibitem{russakovsky2015}
O.~Russakovsky, J.~Deng, H.~Su, J.~Krause, S.~Satheesh, S.~Ma, Z.~Huang,
  A.~Karpathy, A.~Khosla, M.~Bernstein, A.~C. Berg, L.~Fei-Fei, {ImageNet}
  large scale visual recognition challenge, Int. J. Comput. Vis. 115~(3) (2015)
  211--252.
\newblock \href {https://doi.org/10.1007/s11263-015-0816-y}
  {\path{doi:10.1007/s11263-015-0816-y}}.

\bibitem{richter2016}
S.~R. Richter, V.~Vineet, S.~Roth, V.~Koltun, Playing for data: Ground truth
  from computer games, in: Computer Vision -- ECCV 2016, Vol. 9906 of Lecture
  Notes in Computer Science, 2016, pp. 102--118.
\newblock \href {https://doi.org/10.1007/978-3-319-46475-6_7}
  {\path{doi:10.1007/978-3-319-46475-6_7}}.

\bibitem{bird2020}
J.~J. Bird, D.~R. Faria, A.~Ek{\'a}rt, P.~P.~S. Ayrosa, From simulation to
  reality: {CNN} transfer learning for scene classification, in: 2020 IEEE 10th
  International Conference on Intelligent Systems (IS), 2020, pp. 619--625.
\newblock \href {https://doi.org/10.1109/IS48319.2020.9199968}
  {\path{doi:10.1109/IS48319.2020.9199968}}.

\bibitem{simonyan2015}
K.~Simonyan, A.~Zisserman, \href{https://arxiv.org/abs/1409.1556}{Very deep
  convolutional networks for large-scale image recognition}, in: International
  Conference on Learning Representations (ICLR), 2015, arXiv:1409.1556.
\newline\urlprefix\url{https://arxiv.org/abs/1409.1556}

\bibitem{muralidharan2026}
S.~Muralidharan, H.~R. Medeiros, M.~Aminbeidokhti, E.~Granger, M.~Pedersoli,
  Pretraining helps when capacity allows: Evidence from ultra-small {ConvNets},
  in: 2026 IEEE/CVF Winter Conference on Applications of Computer Vision
  (WACV), 2026, pp. 8333--8342.
\newblock \href {https://doi.org/10.1109/WACV61042.2026.00804}
  {\path{doi:10.1109/WACV61042.2026.00804}}.

\bibitem{jacob2018}
B.~Jacob, S.~Kligys, B.~Chen, M.~Zhu, M.~Tang, A.~Howard, H.~Adam,
  D.~Kalenichenko, Quantization and training of neural networks for efficient
  integer-arithmetic-only inference, in: 2018 IEEE/CVF Conference on Computer
  Vision and Pattern Recognition (CVPR), 2018, pp. 2704--2713.
\newblock \href {https://doi.org/10.1109/CVPR.2018.00286}
  {\path{doi:10.1109/CVPR.2018.00286}}.

\bibitem{krishnamoorthi2018}
R.~Krishnamoorthi, Quantizing deep convolutional networks for efficient
  inference: A whitepaper, arXiv preprint, arXiv:1806.08342 (2018).
\newblock \href {https://doi.org/10.48550/arXiv.1806.08342}
  {\path{doi:10.48550/arXiv.1806.08342}}.

\bibitem{sandler2018}
M.~Sandler, A.~Howard, M.~Zhu, A.~Zhmoginov, L.-C. Chen, {MobileNetV2}:
  Inverted residuals and linear bottlenecks, in: 2018 IEEE/CVF Conference on
  Computer Vision and Pattern Recognition (CVPR), 2018, pp. 4510--4520.
\newblock \href {https://doi.org/10.1109/CVPR.2018.00474}
  {\path{doi:10.1109/CVPR.2018.00474}}.

\bibitem{kataoka2021}
H.~Kataoka, K.~Okayasu, A.~Matsumoto, E.~Yamagata, R.~Yamada, N.~Inoue,
  A.~Nakamura, Y.~Satoh, Pre-training without natural images, in: Computer
  Vision -- ACCV 2020, Vol. 12627 of Lecture Notes in Computer Science, 2021,
  pp. 583--600.
\newblock \href {https://doi.org/10.1007/978-3-030-69544-6_35}
  {\path{doi:10.1007/978-3-030-69544-6_35}}.

\bibitem{borkman2021}
S.~Borkman, A.~Crespi, S.~Dhakad, S.~Ganguly, J.~Hogins, Y.-C. Jhang,
  M.~Kamalzadeh, B.~Li, S.~Leal, P.~Parisi, C.~Romero, W.~Smith, A.~Thaman,
  S.~Warren, N.~Yadav, Unity {Perception}: Generate synthetic data for computer
  vision, arXiv preprint, arXiv:2107.04259 (2021).
\newblock \href {https://doi.org/10.48550/arXiv.2107.04259}
  {\path{doi:10.48550/arXiv.2107.04259}}.

\bibitem{lin2020}
J.~Lin, W.-M. Chen, Y.~Lin, J.~Cohn, C.~Gan, S.~Han,
  \href{https://papers.nips.cc/paper/2020/hash/86c51678350f656dcc7f490a43946ee5-Abstract.html}{{MCUNet}:
  Tiny deep learning on {IoT} devices}, in: Advances in Neural Information
  Processing Systems, Vol.~33, 2020, pp. 11711--11722.
\newline\urlprefix\url{https://papers.nips.cc/paper/2020/hash/86c51678350f656dcc7f490a43946ee5-Abstract.html}

\bibitem{bouthillier2021}
X.~Bouthillier, P.~Delaunay, M.~Bronzi, A.~Trofimov, B.~Nichyporuk, J.~Szeto,
  N.~Mohammadi~Sepahvand, E.~Raff, K.~Madan, V.~Voleti, S.~Ebrahimi~Kahou,
  V.~Michalski, T.~Arbel, C.~Pal, G.~Varoquaux, P.~Vincent,
  \href{https://proceedings.mlsys.org/paper_files/paper/2021/hash/0184b0cd3cfb185989f858a1d9f5c1eb-Abstract.html}{Accounting
  for variance in machine learning benchmarks}, in: Proceedings of Machine
  Learning and Systems, Vol.~3, 2021, pp. 747--769.
\newline\urlprefix\url{https://proceedings.mlsys.org/paper_files/paper/2021/hash/0184b0cd3cfb185989f858a1d9f5c1eb-Abstract.html}

\bibitem{he2019}
K.~He, R.~Girshick, P.~Doll{\'a}r, Rethinking {ImageNet} pre-training, in: 2019
  IEEE/CVF International Conference on Computer Vision (ICCV), 2019, pp.
  4917--4926.
\newblock \href {https://doi.org/10.1109/ICCV.2019.00502}
  {\path{doi:10.1109/ICCV.2019.00502}}.

\bibitem{kornblith2019}
S.~Kornblith, J.~Shlens, Q.~V. Le, Do better {ImageNet} models transfer
  better?, in: 2019 IEEE/CVF Conference on Computer Vision and Pattern
  Recognition (CVPR), 2019, pp. 2656--2666.
\newblock \href {https://doi.org/10.1109/CVPR.2019.00277}
  {\path{doi:10.1109/CVPR.2019.00277}}.

\bibitem{pan2010}
S.~J. Pan, Q.~Yang, A survey on transfer learning, IEEE Trans. Knowl. Data Eng.
  22~(10) (2010) 1345--1359.
\newblock \href {https://doi.org/10.1109/TKDE.2009.191}
  {\path{doi:10.1109/TKDE.2009.191}}.

\bibitem{zhang2023}
W.~Zhang, L.~Deng, L.~Zhang, D.~Wu, A survey on negative transfer, IEEE/CAA J.
  Autom. Sin. 10~(2) (2023) 305--329.
\newblock \href {https://doi.org/10.1109/JAS.2022.106004}
  {\path{doi:10.1109/JAS.2022.106004}}.

\bibitem{catalantatjer2026}
A.~Catalan-Tatjer, N.~Ajroldi, J.~Geiping,
  \href{https://arxiv.org/abs/2510.06213}{Training dynamics impact
  post-training quantization robustness}, in: International Conference on
  Learning Representations (ICLR), 2026, arXiv:2510.06213.
\newline\urlprefix\url{https://arxiv.org/abs/2510.06213}

\bibitem{david2021}
R.~David, J.~Duke, A.~Jain, V.~Janapa~Reddi, N.~Jeffries, J.~Li, N.~Kreeger,
  I.~Nappier, M.~Natraj, S.~Regev, R.~Rhodes, T.~Wang, P.~Warden,
  \href{https://proceedings.mlsys.org/paper_files/paper/2021/hash/6c44dc73014d66ba49b28d483a8f8b0d-Abstract.html}{{TensorFlow
  Lite Micro}: Embedded machine learning for {TinyML} systems}, in: Proceedings
  of Machine Learning and Systems, Vol.~3, 2021, pp. 800--811.
\newline\urlprefix\url{https://proceedings.mlsys.org/paper_files/paper/2021/hash/6c44dc73014d66ba49b28d483a8f8b0d-Abstract.html}

\bibitem{lin2021mcunetv2}
J.~Lin, W.-M. Chen, H.~Cai, C.~Gan, S.~Han,
  \href{https://proceedings.neurips.cc/paper/2021/hash/1371bccec2447b5aa6d96d2a540fb401-Abstract.html}{Memory-efficient
  patch-based inference for tiny deep learning}, in: Advances in Neural
  Information Processing Systems, Vol.~34, 2021, pp. 2346--2358, also
  arXiv:2110.15352 (MCUNetV2).
\newline\urlprefix\url{https://proceedings.neurips.cc/paper/2021/hash/1371bccec2447b5aa6d96d2a540fb401-Abstract.html}

\bibitem{banbury2021micronets}
C.~Banbury, C.~Zhou, I.~Fedorov, R.~Matas, U.~Thakker, D.~Gope,
  V.~Janapa~Reddi, M.~Mattina, P.~Whatmough,
  \href{https://proceedings.mlsys.org/paper_files/paper/2021/hash/c4d41d9619462c534b7b61d1f772385e-Abstract.html}{{MicroNets}:
  Neural network architectures for deploying {TinyML} applications on commodity
  microcontrollers}, in: Proceedings of Machine Learning and Systems, Vol.~3,
  2021, pp. 517--532.
\newline\urlprefix\url{https://proceedings.mlsys.org/paper_files/paper/2021/hash/c4d41d9619462c534b7b61d1f772385e-Abstract.html}

\bibitem{wilson2026}
B.~Wilson, {TinySSL}: Distilled self-supervised pretraining for sub-megabyte
  {MCU} models, arXiv preprint, arXiv:2605.08241 (2026).
\newblock \href {https://doi.org/10.48550/arXiv.2605.08241}
  {\path{doi:10.48550/arXiv.2605.08241}}.

\bibitem{faggi2026}
D.~Faggi, O.~Kuznetsov, S.~Mereu, A.~Galdelli, E.~Frontoni, M.~Arnesano, Hybrid
  {CNN-KAN} architectures for ultra-compact person detection on {IoT} devices:
  A systematic evaluation of quantization strategies, IEEE Access 14 (2026)
  81696--81712.
\newblock \href {https://doi.org/10.1109/ACCESS.2026.3693881}
  {\path{doi:10.1109/ACCESS.2026.3693881}}.

\bibitem{espnn2026}
{Espressif Systems},
  \href{https://components.espressif.com/components/espressif/esp-nn/versions/1.3.2/readme}{{ESP-NN}:
  Optimised neural network functions for {Espressif} chips (software, version
  1.3.2)}, ESP Component Registry, repository commit
  c51d5fa5ecbd3ed7200078a5fc9c1f5a8a077d36; accessed 1 October 2026 (2026).
\newline\urlprefix\url{https://components.espressif.com/components/espressif/esp-nn/versions/1.3.2/readme}

\bibitem{zhang2025}
K.~Zhang, H.~Poonawala, Optimizing {E}dge-{AI} on the {ESP32-S3} using {SIMD},
  in: 2025 IEEE International Performance, Computing, and Communications
  Conference (IPCCC), Austin, TX, USA, 2025, pp. 1--2.
\newblock \href {https://doi.org/10.1109/IPCCC66453.2025.11304640}
  {\path{doi:10.1109/IPCCC66453.2025.11304640}}.

\bibitem{tflmoptkernels}
{TensorFlow Lite Micro authors},
  \href{https://github.com/tensorflow/tflite-micro/blob/main/tensorflow/lite/micro/docs/optimized_kernel_implementations.md}{Optimized
  kernel implementations}, Documentation in the tflite-micro repository,
  \texttt{tensorflow/lite/micro/docs/optimized\_kernel\_implementations.md},
  read in the copy distributed with Zephyr's tflite-micro module (3.4.0-2.7.6);
  accessed 24 October 2026 (2026).
\newline\urlprefix\url{https://github.com/tensorflow/tflite-micro/blob/main/tensorflow/lite/micro/docs/optimized_kernel_implementations.md}

\bibitem{schlogl2023}
A.~Schl{\"o}gl, N.~Hofer, R.~B{\"o}hme, Causes and effects of unanticipated
  numerical deviations in neural network inference frameworks, in: Advances in
  Neural Information Processing Systems, Vol.~36, 2023, pp. 56095--56107.
\newblock \href {https://doi.org/10.52202/075280-2446}
  {\path{doi:10.52202/075280-2446}}.

\bibitem{yadav2026osf}
A.~Yadav, \href{https://osf.io/bpqga}{Synthetic-to-real transfer learning for
  scene classification on microcontrollers: pre-registered analysis plan}, OSF
  Registries, registered 13 August 2026; accessed 1 October 2026 (2026).
\newblock \href {https://doi.org/10.17605/OSF.IO/BPQGA}
  {\path{doi:10.17605/OSF.IO/BPQGA}}.
\newline\urlprefix\url{https://osf.io/bpqga}

\bibitem{nosek2018}
B.~A. Nosek, C.~R. Ebersole, A.~C. DeHaven, D.~T. Mellor, The preregistration
  revolution, Proc. Natl. Acad. Sci. U. S. A. 115~(11) (2018) 2600--2606.
\newblock \href {https://doi.org/10.1073/pnas.1708274114}
  {\path{doi:10.1073/pnas.1708274114}}.

\bibitem{willroth2024}
E.~C. Willroth, O.~E. Atherton, Best laid plans: A guide to reporting
  preregistration deviations, Adv. Methods Pract. Psychol. Sci. 7~(1) (2024).
\newblock \href {https://doi.org/10.1177/25152459231213802}
  {\path{doi:10.1177/25152459231213802}}.

\bibitem{yadav2026osfaddendum}
A.~Yadav, \href{https://osf.io/cb7x3}{Addendum to the registration of 13
  {A}ugust 2026: exploratory work after the registered analysis}, OSF
  Registries, open-ended registration filed 5 October 2026; names the
  registration of 13 August 2026 and leaves it unchanged; accessed 24 October
  2026 (2026).
\newblock \href {https://doi.org/10.17605/OSF.IO/CB7X3}
  {\path{doi:10.17605/OSF.IO/CB7X3}}.
\newline\urlprefix\url{https://osf.io/cb7x3}

\bibitem{bird2020data}
J.~J. Bird,
  \href{https://www.kaggle.com/datasets/birdy654/environment-recognition-simulation-to-reality/versions/1}{[dataset]
  {S}cene classification: Simulation to reality}, Kaggle, version 1, released 2
  January 2020; accessed 1 October 2026 (2020).
\newline\urlprefix\url{https://www.kaggle.com/datasets/birdy654/environment-recognition-simulation-to-reality/versions/1}

\bibitem{imagehash}
J.~Buchner, \href{https://pypi.org/project/ImageHash/4.3.2/}{{ImageHash}: A
  {Python} perceptual image hashing module (software, version 4.3.2)}, Python
  Package Index, released 1 February 2025; source at
  https://github.com/JohannesBuchner/imagehash; accessed 1 October 2026 (2025).
\newline\urlprefix\url{https://pypi.org/project/ImageHash/4.3.2/}

\bibitem{pedregosa2011}
F.~Pedregosa, G.~Varoquaux, A.~Gramfort, V.~Michel, B.~Thirion, O.~Grisel,
  M.~Blondel, P.~Prettenhofer, R.~Weiss, V.~Dubourg, J.~Vanderplas, A.~Passos,
  D.~Cournapeau, M.~Brucher, M.~Perrot, {\'E}.~Duchesnay,
  \href{https://www.jmlr.org/papers/v12/pedregosa11a.html}{Scikit-learn:
  Machine learning in {Python}}, J. Mach. Learn. Res. 12~(85) (2011)
  2825--2830, accessed 1 October 2026.
\newline\urlprefix\url{https://www.jmlr.org/papers/v12/pedregosa11a.html}

\bibitem{chollet2015keras}
F.~Chollet, et~al., {Keras} (software, version 3.13.2), \url{https://keras.io},
  accessed 1 October 2026 (2026).

\bibitem{howard2019imagenette}
J.~Howard, \href{https://github.com/fastai/imagenette}{[dataset] {Imagenette}:
  A subset of 10 easily classified classes from {ImageNet}}, GitHub, version 2
  split, 160\,px; loaded via TensorFlow Datasets (imagenette/160px-v2, 1.0.0);
  accessed 1 October 2026 (2019).
\newline\urlprefix\url{https://github.com/fastai/imagenette}

\bibitem{tensorflow2015}
M.~Abadi, A.~Agarwal, P.~Barham, E.~Brevdo, Z.~Chen, C.~Citro, G.~S. Corrado,
  A.~Davis, J.~Dean, M.~Devin, S.~Ghemawat, I.~Goodfellow, A.~Harp, G.~Irving,
  M.~Isard, Y.~Jia, R.~Jozefowicz, L.~Kaiser, M.~Kudlur, J.~Levenberg,
  D.~Man{\'e}, R.~Monga, S.~Moore, D.~Murray, C.~Olah, M.~Schuster, J.~Shlens,
  B.~Steiner, I.~Sutskever, K.~Talwar, P.~Tucker, V.~Vanhoucke, V.~Vasudevan,
  F.~Vi{\'e}gas, O.~Vinyals, P.~Warden, M.~Wattenberg, M.~Wicke, Y.~Yu,
  X.~Zheng, {TensorFlow}: Large-scale machine learning on heterogeneous systems
  (software, version 2.20.0), \url{https://www.tensorflow.org}, software
  available from tensorflow.org; accessed 1 October 2026 (2025).
\newblock \href {https://doi.org/10.5281/zenodo.16852354}
  {\path{doi:10.5281/zenodo.16852354}}.

\bibitem{kingma2015}
D.~P. Kingma, J.~Ba, \href{https://arxiv.org/abs/1412.6980}{Adam: A method for
  stochastic optimization}, in: International Conference on Learning
  Representations (ICLR), 2015, arXiv:1412.6980.
\newline\urlprefix\url{https://arxiv.org/abs/1412.6980}

\bibitem{litertptq}
{Google AI Edge},
  \href{https://developers.google.com/edge/litert/conversion/tensorflow/quantization/post_training_integer_quant}{Post-training
  integer quantization}, LiteRT documentation, last updated 28 May 2026;
  accessed 1 October 2026 (2026).
\newline\urlprefix\url{https://developers.google.com/edge/litert/conversion/tensorflow/quantization/post_training_integer_quant}

\bibitem{nagel2021}
M.~Nagel, M.~Fournarakis, R.~A. Amjad, Y.~Bondarenko, M.~van Baalen,
  T.~Blankevoort, A white paper on neural network quantization, arXiv preprint,
  arXiv:2106.08295 (2021).
\newblock \href {https://doi.org/10.48550/arXiv.2106.08295}
  {\path{doi:10.48550/arXiv.2106.08295}}.

\bibitem{dietterich1998}
T.~G. Dietterich, Approximate statistical tests for comparing supervised
  classification learning algorithms, Neural Comput. 10~(7) (1998) 1895--1923.
\newblock \href {https://doi.org/10.1162/089976698300017197}
  {\path{doi:10.1162/089976698300017197}}.

\bibitem{nadeau2003}
C.~Nadeau, Y.~Bengio, Inference for the generalization error, Mach. Learn.
  52~(3) (2003) 239--281.
\newblock \href {https://doi.org/10.1023/A:1024068626366}
  {\path{doi:10.1023/A:1024068626366}}.

\bibitem{demsar2006}
J.~Dem{\v{s}}ar,
  \href{https://www.jmlr.org/papers/v7/demsar06a.html}{Statistical comparisons
  of classifiers over multiple data sets}, J. Mach. Learn. Res. 7~(1) (2006)
  1--30, accessed 1 October 2026.
\newline\urlprefix\url{https://www.jmlr.org/papers/v7/demsar06a.html}

\bibitem{holm1979}
S.~Holm, \href{https://www.jstor.org/stable/4615733}{A simple sequentially
  rejective multiple test procedure}, Scand. J. Stat. 6~(2) (1979) 65--70.
\newline\urlprefix\url{https://www.jstor.org/stable/4615733}

\bibitem{espidf602}
{Espressif Systems},
  \href{https://github.com/espressif/esp-idf/releases/tag/v6.0.2}{{ESP-IDF}:
  Espressif {IoT} development framework (software, version 6.0.2)}, GitHub,
  released 29 June 2026; accessed 2 October 2026 (2026).
\newline\urlprefix\url{https://github.com/espressif/esp-idf/releases/tag/v6.0.2}

\bibitem{esptflitemicro140}
{Espressif Systems},
  \href{https://components.espressif.com/components/espressif/esp-tflite-micro/versions/1.4.0}{{esp-tflite-micro}:
  {TensorFlow Lite Micro} component for {ESP-IDF} (software, version 1.4.0)},
  ESP Component Registry, version used; not redistributed; accessed 2 October
  2026 (2026).
\newline\urlprefix\url{https://components.espressif.com/components/espressif/esp-tflite-micro/versions/1.4.0}

\bibitem{espnn131}
{Espressif Systems},
  \href{https://components.espressif.com/components/espressif/esp-nn/versions/1.3.1/readme}{{ESP-NN}:
  Optimised neural network functions for {Espressif} chips (software, version
  1.3.1)}, ESP Component Registry, used with a local patch to the tiled
  depthwise kernel; the patch is available from the author on request; accessed
  2 October 2026 (2026).
\newline\urlprefix\url{https://components.espressif.com/components/espressif/esp-nn/versions/1.3.1/readme}

\bibitem{esptflm140planner}
{Espressif Systems},
  \href{https://github.com/espressif/esp-tflite-micro/tree/v1.4.0}{Greedy
  memory planner header, \texttt{greedy\_memory\_planner.h}, in
  {esp-tflite-micro} (software, version 1.4.0)}, GitHub, comment at lines
  110--113; tag commit 75e189e3; accessed 1 November 2026 (2026).
\newline\urlprefix\url{https://github.com/espressif/esp-tflite-micro/tree/v1.4.0}

\bibitem{tflmissue2290}
{TensorFlow Lite Micro contributors},
  \href{https://github.com/tensorflow/tflite-micro/issues/2290}{Issue \#2290,
  tensorflow/tflite-micro}, GitHub, contributor reply of 27 October 2023;
  accessed 1 November 2026 (2023).
\newline\urlprefix\url{https://github.com/tensorflow/tflite-micro/issues/2290}

\bibitem{arduinopico560}
{arduino-pico contributors},
  \href{https://github.com/earlephilhower/arduino-pico/releases/tag/5.6.0}{{arduino-pico}:
  {Raspberry Pi Pico} {Arduino} core for {RP2040} and {RP2350} boards
  (software, version 5.6.0)}, GitHub, released 26 April 2026; accessed 2
  October 2026 (2026).
\newline\urlprefix\url{https://github.com/earlephilhower/arduino-pico/releases/tag/5.6.0}

\bibitem{microtflite104}
J.~O'Sullivan, {TensorFlow Authors},
  \href{https://github.com/johnosbb/MicroTFLite}{{MicroTFLite}: An
  {Arduino}-friendly version of {TensorFlow Lite} (software, version 1.0.4)},
  GitHub, version used; not redistributed; accessed 2 October 2026 (2025).
\newline\urlprefix\url{https://github.com/johnosbb/MicroTFLite}

\bibitem{ppk2}
{Nordic Semiconductor},
  \href{https://www.nordicsemi.com/Products/Development-hardware/Power-Profiler-Kit-2}{{Power
  Profiler Kit II}}, Product page, accessed 2 October 2026 (2026).
\newline\urlprefix\url{https://www.nordicsemi.com/Products/Development-hardware/Power-Profiler-Kit-2}

\bibitem{rpihwdesign}
{Raspberry Pi Ltd},
  \href{https://datasheets.raspberrypi.com/rp2040/hardware-design-with-rp2040.pdf}{Hardware
  design with {RP2040}: Using {RP2040} microcontrollers to build boards and
  products}, Datasheet RP-008279-DS-2, Release 2, build date 20 August 2026;
  see Sec. 3.1.1; accessed 2 October 2026 (2026).
\newline\urlprefix\url{https://datasheets.raspberrypi.com/rp2040/hardware-design-with-rp2040.pdf}

\bibitem{picodatasheet}
{Raspberry Pi Ltd},
  \href{https://datasheets.raspberrypi.com/pico/pico-datasheet.pdf}{{Raspberry
  Pi Pico} datasheet: An {RP2040}-based microcontroller board}, Datasheet,
  Release 21, build date 3 July 2026; see Secs. 2.1 and 4.5; accessed 25
  September 2026 (2026).
\newline\urlprefix\url{https://datasheets.raspberrypi.com/pico/pico-datasheet.pdf}

\bibitem{mlperftiny}
C.~Banbury, V.~Janapa~Reddi, P.~Torelli, N.~Jeffries, C.~Kiraly, J.~Holleman,
  P.~Montino, D.~Kanter, P.~Warden, D.~Pau, U.~Thakker, A.~Torrini, J.~Cordaro,
  G.~Di~Guglielmo, J.~Duarte, H.~Tran, N.~Tran, W.~Niu, X.~Xu,
  \href{https://datasets-benchmarks-proceedings.neurips.cc/paper/2021/hash/da4fb5c6e93e74d3df8527599fa62642-Abstract-round1.html}{{MLPerf
  Tiny} benchmark}, in: Proceedings of the Neural Information Processing
  Systems Track on Datasets and Benchmarks, Vol.~1, 2021, also
  arXiv:2106.07597; accessed 2 October 2026.
\newline\urlprefix\url{https://datasets-benchmarks-proceedings.neurips.cc/paper/2021/hash/da4fb5c6e93e74d3df8527599fa62642-Abstract-round1.html}

\bibitem{espnn141}
{Espressif Systems},
  \href{https://components.espressif.com/components/espressif/esp-nn/versions/1.4.1/readme}{{ESP-NN}:
  Optimised neural network functions for {Espressif} chips (software, version
  1.4.1)}, ESP Component Registry, repository commit
  d9be5422a79f4115e4275452dcaba3fe363a697d; used unpatched; accessed 24 October
  2026 (2026).
\newline\urlprefix\url{https://components.espressif.com/components/espressif/esp-nn/versions/1.4.1/readme}

\bibitem{espnnpr42}
A.~Yadav, \href{https://github.com/espressif/esp-nn/pull/42}{fix(max\_pool/s3):
  apply activation\_min when channels \% 8 == 4}, GitHub pull request \#42,
  espressif/esp-nn, merged into master on 8 October 2026 (merge commit
  c972c32a); accessed 24 October 2026 (2026).
\newline\urlprefix\url{https://github.com/espressif/esp-nn/pull/42}

\bibitem{esptflitemicro141}
{Espressif Systems},
  \href{https://components.espressif.com/components/espressif/esp-tflite-micro/versions/1.4.1/readme}{{esp-tflite-micro}:
  {TensorFlow Lite Micro} component for {ESP-IDF} (software, version 1.4.1)},
  ESP Component Registry, repository commit
  5d505b5ffac8e60923bf29a0cc6e6348c84e3763; accessed 24 October 2026 (2026).
\newline\urlprefix\url{https://components.espressif.com/components/espressif/esp-tflite-micro/versions/1.4.1/readme}

\bibitem{tobin2017}
J.~Tobin, R.~Fong, A.~Ray, J.~Schneider, W.~Zaremba, P.~Abbeel, Domain
  randomization for transferring deep neural networks from simulation to the
  real world, in: 2017 IEEE/RSJ International Conference on Intelligent Robots
  and Systems (IROS), 2017, pp. 23--30.
\newblock \href {https://doi.org/10.1109/IROS.2017.8202133}
  {\path{doi:10.1109/IROS.2017.8202133}}.

\bibitem{baradad2021}
M.~Baradad, J.~Wulff, T.~Wang, P.~Isola, A.~Torralba,
  \href{https://arxiv.org/abs/2106.05963}{Learning to see by looking at noise},
  in: Advances in Neural Information Processing Systems, Vol.~34, 2021, pp.
  2556--2569, also arXiv:2106.05963.
\newline\urlprefix\url{https://arxiv.org/abs/2106.05963}

\bibitem{mishra2022}
S.~Mishra, R.~Panda, C.~P. Phoo, C.-F.~R. Chen, L.~Karlinsky, K.~Saenko,
  V.~Saligrama, R.~S. Feris, {Task2Sim}: Towards effective pre-training and
  transfer from synthetic data, in: 2022 IEEE/CVF Conference on Computer Vision
  and Pattern Recognition (CVPR), 2022, pp. 9194--9204.
\newblock \href {https://doi.org/10.1109/CVPR52688.2022.00898}
  {\path{doi:10.1109/CVPR52688.2022.00898}}.

\end{thebibliography}

\end{document}